\pdfoutput=1
\documentclass[lettersize,journal]{IEEEtran}

\PassOptionsToPackage{hyphens}{url}
\PassOptionsToPackage{bookmarks=false}{hyperref}

\newcommand{\IEEEauthororcidlink}[1]{}

\usepackage[utf8]{inputenc}
\usepackage[T1]{fontenc}

\usepackage{xurl}

\usepackage{soul}
\usepackage[dvipsnames,table]{xcolor}
\sethlcolor{yellow!45}
\soulregister\textsc7
\soulregister\textbf7
\soulregister\emph7
\soulregister\ref7
\soulregister\cite7
\soulregister\texttt7
\soulregister\footnote7
\usepackage{microtype}

\usepackage{dblfloatfix}

\usepackage{amsmath,amssymb,mathtools}
\usepackage{bm}
\usepackage{booktabs}
\usepackage{multirow}
\usepackage{array}
\usepackage{siunitx}
\usepackage{threeparttable}
\usepackage{threeparttablex}
\usepackage{tabularx}
\usepackage{makecell}
\usepackage{pbox}
\usepackage{algorithm}
\usepackage{algorithmic}

\newcolumntype{L}[1]{>{\raggedright\arraybackslash}p{#1}}
\newcolumntype{C}[1]{>{\centering\arraybackslash}p{#1}}
\newcolumntype{R}[1]{>{\raggedleft\arraybackslash}p{#1}}
\newcolumntype{Y}{>{\raggedright\arraybackslash}X}

\PassOptionsToPackage{final}{graphicx}   
\usepackage{graphicx}
\setkeys{Gin}{draft=false}

\graphicspath{{./}{./figures/}}
\DeclareGraphicsExtensions{.pdf,.png,.jpg,.jpeg}
\usepackage[caption=false,font=footnotesize]{subfig}

\usepackage{placeins}
\usepackage{balance}

\usepackage{cite}
\usepackage{url}

\usepackage{hyperref}

\hypersetup{
    colorlinks=true,
    linkcolor=black,
    citecolor=NavyBlue,
    urlcolor=black
}

\usepackage[capitalise,noabbrev,nameinlink]{cleveref}

\DeclareSIUnit\rpm{rpm}
\DeclareSIUnit\db{dB}
\DeclareSIUnit\fps{fps}
\DeclareSIUnit\ms{ms}

\begin{document}

\title{Reliability-Calibrated Edge-IoT Early Fault Warning for Rotating Machinery with a Physics-Guided Tiny-Mamba Transformer}

\author{%
  Changyu~Li$^{1}$\IEEEauthororcidlink{0009-0007-0876-0310},
  Huabei~Nie$^{2}$\IEEEauthororcidlink{0009-0007-4996-9989},
  Xiaoya~Ni$^{3}$\IEEEauthororcidlink{0009-0007-4789-2906},
  Lu~Wang$^{4}$\IEEEauthororcidlink{0000-0001-6345-3873},~\IEEEmembership{Senior Member,~IEEE},
  Lijuan~Shen$^{5}$\IEEEauthororcidlink{0000-0002-3906-4250},
  Kaishun~Wu$^{6}$\IEEEauthororcidlink{0000-0003-2216-0737},~\IEEEmembership{Fellow,~IEEE},
  and Fei~Luo$^{1}$\IEEEauthororcidlink{0000-0001-9760-1520}%

  \thanks{This work has been submitted to the IEEE Internet of Things Journal for possible publication. This arXiv version is an author-submitted preprint.}%
  \thanks{$^{1}$Changyu~Li and Fei~Luo are with Great Bay University, Dongguan, China (e-mail: Changyuli.021230@gmail.com; luofei2018@outlook.com).}%
  \thanks{$^{2}$Huabei~Nie is with the School of Computer Science and Engineering, Huizhou University, Huizhou 516007, China (e-mail: 331096487@qq.com).}%
  \thanks{$^{3}$Xiaoya~Ni is with the National University of Singapore, Singapore 119077 (e-mail: e1520186@u.nus.edu).}%
  \thanks{$^{4}$Lu~Wang is with the College of Computer Science and Software Engineering, Shenzhen University, Shenzhen, China (e-mail: wanglu@szu.edu.cn).}%
  \thanks{$^{5}$Lijuan~Shen is with James Cook University, Singapore 387380 (e-mail: lijuan.shen@jcu.edu.au).}%
  \thanks{$^{6}$Kaishun~Wu is with the Hong Kong University of Science and Technology (Guangzhou), Guangzhou, China (e-mail: wuks@hkust-gz.edu.cn).}%
}

\markboth{Author-submitted preprint}%
{Li \MakeLowercase{\textit{et al.}}: Reliability-Calibrated Edge-IoT Early Fault Warning with a Physics-Guided Tiny-Mamba Transformer}

\maketitle

\begin{abstract}
Industrial Internet of Things (IIoT) systems increasingly rely on distributed vibration sensing to support predictive maintenance of rotating machinery. In practical deployments, however, raw signal upload is costly and alarm decisions must be made locally under limited computation, changing operating conditions, and strict nuisance-alarm budgets. This paper presents a reliability-calibrated edge-IoT early-warning framework in which a compact Physics-Guided Tiny-Mamba Transformer (PG-TMT) acts as the representation module and an extreme value theory (EVT) layer converts streaming anomaly scores into event-level alarm episodes. PG-TMT combines a depthwise-separable convolutional stem, a Tiny-Mamba state-space branch, and a lightweight local Transformer to capture transient, long-horizon, and multichannel degradation cues under \texttt{batch}=1 inference. To improve auditability, temporal attention is projected to the frequency domain and softly aligned with analytical bearing fault-order bands. EVT calibration, dual-threshold hysteresis, and trimmed-tail fitting then provide controllable false-alarm intensity even when healthy calibration data are imperfect. Experiments on CWRU, Paderborn, XJTU-SY, and an industrial pilot show that the framework improves PR--AUC, reduces detection delay at a controlled nuisance budget, and remains robust to structured interference, metadata uncertainty, compound fault mixtures, and domain transfer. With a sub-1\,MB footprint and Jetson p99 latency below 7\,ms, the framework supports calibrated and interpretable early warnings for IIoT predictive maintenance.
\end{abstract}

\begin{IEEEkeywords}
Industrial Internet of Things (IIoT), edge intelligence, predictive maintenance, rotating machinery, early fault warning, reliability calibration, streaming anomaly detection, extreme value theory (EVT), physics-guided learning, Tiny-Mamba, Transformers.
\end{IEEEkeywords}

\section{Introduction}
\begin{figure*}[!t]
  \centering
  \includegraphics[width=0.92\textwidth]{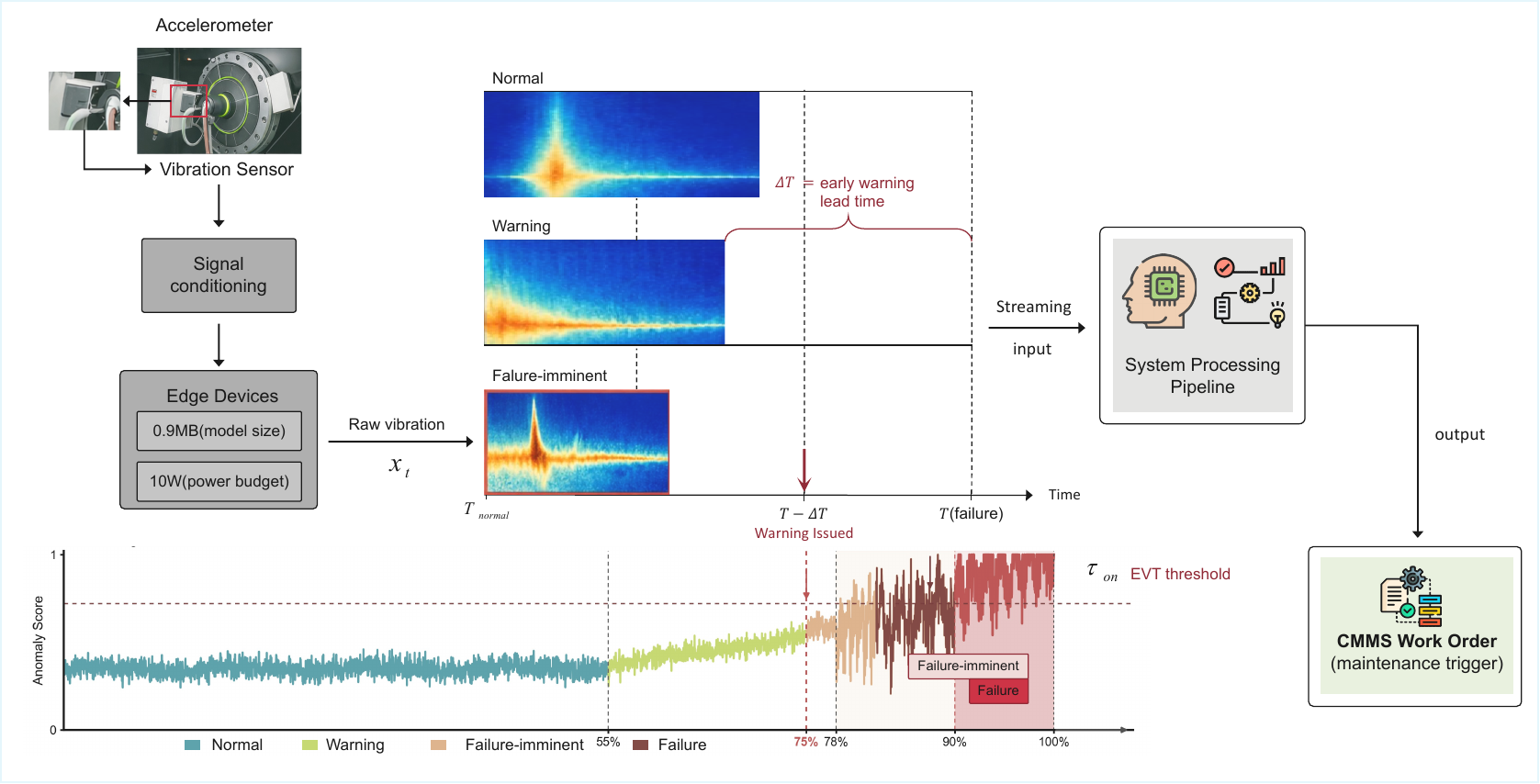}
  \caption{\textsc{PG-TMT} edge-IoT early warning workflow and deployment interface. High-rate vibration streams are collected from rotating machinery, conditioned at the edge, and processed locally by PG-TMT with \texttt{batch}=1 streaming inference. The system converts streaming anomaly scores into calibrated alarm episodes through EVT thresholding and dual-threshold hysteresis, enabling warning before failure. In deployment, operating metadata such as RPM and bearing geometry are used to update the physics prior, while alarm events and score summaries are logged for inspection, CMMS work orders, and reliability verification.}
  \label{fig:industry_flow}
\end{figure*}

\IEEEPARstart{T}{he} Internet of Things (IoT) is becoming part of many industrial systems by connecting sensors, machines, edge devices, and maintenance services. In the Industrial Internet of Things (IIoT), condition monitoring is moving from periodic inspection and cloud analysis to continuous sensing and edge intelligence. Modern production lines contain many rotating assets, such as bearings, gears, motors, pumps, and fans. Their vibration streams contain useful information about normal operation, operating changes, and early degradation. Processing these streams close to the machine is important, since cloud monitoring may introduce communication overhead, latency, privacy exposure, and dependence on network availability. Edge computing therefore offers a practical path for low-latency industrial monitoring, where raw vibration data remain local and only compact scores or alarm events are transmitted~\cite{Sisinni2018IIoTChallenges,Shi2016EdgeComputing}. Fig.~\ref{fig:industry_flow} illustrates the proposed edge-IoT workflow from vibration sensing to maintenance action.

Early fault warning for rotating machinery is a typical IIoT application. Bearings and gears often run for long periods under changing speed and load, and they are usually part of tightly coupled production systems. In such settings, a small change in detection delay or false-alarm frequency can make a large difference in uptime, maintenance cost, and downstream risk~\cite{Lei2016MSSPReview,Zhu2023MeasurementReview}. Unlike offline fault diagnosis, an IIoT warning service must run continuously at \texttt{batch}=1 under limited compute, memory, and energy budgets. It must also handle changing operating conditions, sensor and machine shifts, and severe class imbalance, because healthy operation dominates most industrial streams. Thus, practical alarm policies should be accurate, calibrated, interpretable, and controllable in terms of false-alarm intensity, such as nuisance alarm episodes per hour~\cite{Si2011EJOR}.

Existing rotating machinery monitoring methods can be broadly divided into three categories. The first category is physics-based signal analysis, which uses resonance bands, cyclostationarity, envelope analysis, and defect order harmonics to characterize fault-related vibration patterns. These methods are closely aligned with engineering intuition and field practice~\cite{Randall2011Tutorial,Antoni2007Cyclo,McFadden1984JSV,Yao2019TIM}. The second category is data-driven representation learning. Convolutional networks, recurrent models, Transformers, and state-space models can learn temporal and spectral features directly from waveforms and often achieve strong benchmark accuracy~\cite{Vaswani2017Attention,Xu2022AnomalyTransformer,Gu2023Mamba,Gu2022S4}. Recent studies have further adapted Mamba and SSM backbones to vibration diagnosis through lightweight Mamba models, multi-sensor networks, Transformer-Mamba fusion, and OOD augmentation under zero-faulty data~\cite{Yi2025VibrMamba,Zhang2025GsXANet,Xia2025BMTMNet,Chen2025SMRN}. The third category is physics-guided and knowledge-driven learning, where mechanistic priors are introduced into neural models to improve robustness and interpretability~\cite{Karniadakis2021NRP,Yan2025PNMF,Chi2022IoTJKnowledgeFD,Wang2024IoTJDualDriven}.

Although these methods have advanced rotating machinery diagnosis, several issues remain when they are used as IIoT warning services. First, many deep models are evaluated with static classification metrics, while deployment is a streaming and event based problem. A high offline AUC does not necessarily lead to stable alarm episodes when thresholds are selected heuristically. Second, conventional thresholding provides limited control over false-alarm intensity, which is important for maintenance teams because frequent nuisance alarms can overload inspection resources. Third, physical evidence, learned representations, and statistical calibration are often designed separately, making it difficult to provide both auditable explanations and reliable operating points. Finally, real industrial streams may include structured interference, imperfect speed and geometry metadata, and contaminated ``healthy'' calibration windows. Degradation models, survival analysis, availability modeling, and EVT based anomaly detection provide useful tools for timing evaluation, threshold calibration, and economic assessment~\cite{Si2011EJOR,Siffer2017SPOT,Vignotto2020GPD,Compare2017TR_Avail,Sun2012TR_Benefits}, but they are rarely integrated with compact edge-ready representation learning.

Recent studies in the IEEE Internet of Things Journal have explored real-time bearing diagnosis on resource-constrained devices~\cite{Hu2024IoTJSiameseBearing}, unsupervised bearing anomaly detection from healthy data~\cite{Liu2024IoTJAnoWGAN}, and early fault detection under changing IIoT operating conditions~\cite{Zhao2024IoTJDyEdgeGAT}. These studies show the importance of edge-oriented and IIoT-aware fault monitoring. However, compact representation learning, physical interpretability, alarm intensity calibration, and censoring-aware timing evaluation are still mostly treated as separate components. This motivates a unified framework that supports edge inference, physics-aligned representation, EVT-calibrated false-alarm control, and reliability-calibrated early warning.

In this paper, we propose such a reliability-calibrated edge-IoT framework for rotating machinery early warning. The framework models healthy-score tails with extreme value theory (EVT) to control false-alarm intensity, treats detection as a censorable time-to-event outcome, and connects detection and nuisance alarms to availability and return-on-investment models~\cite{Si2011EJOR,Compare2017TR_Avail,Sun2012TR_Benefits}. As the representation module, we develop the Physics-Guided Tiny-Mamba Transformer (PG-TMT), a compact tri-branch encoder that combines a depthwise-separable convolutional stem, a Tiny-Mamba state-space branch, and a lightweight local-attention branch. To provide interpretable and shift-robust evidence, PG-TMT maps temporal attention into the spectral domain and aligns it with analytical fault-order bands derived from bearing geometry and rotational speed. The resulting scores are converted into stable alarm episodes through EVT calibration, dual-threshold hysteresis, and trimmed-tail fitting for contaminated healthy data.

We evaluate the framework under a leakage-free streaming protocol with sliding windows, \texttt{batch}=1 inference, and healthy-only calibration shared across all methods. We report PR--AUC and ROC--AUC together with censoring-aware timeliness, mean time-to-detect at matched false-alarm intensity, Numenta-style early warning scores, and calibration measures. Robustness is evaluated under additive and structured industrial noise, domain and sensor shifts, metadata uncertainty, proxy compound faults, and contaminated calibration. Experiments are conducted on CWRU, Paderborn, XJTU-SY, and an industrial pilot~\cite{Hendriks2022CWRU,Smith2015CWRUStudy,Lessmeier2016Paderborn,Lei2019XJTUSY_Tutorial}.

The main contributions are summarized as follows.

1) We formulate rotating-machinery early warning as a reliability-calibrated edge-IoT service. Unlike static fault classification, the proposed framework processes high-rate vibration windows locally and produces event-level alarm episodes under a practitioner-specified nuisance-alarm budget.

2) We develop PG-TMT as a compact representation module for \texttt{batch}=1 embedded inference. The convolutional, state-space, and local-attention branches are designed to capture transient impacts, slow degradation trends, and cross-channel resonances within a sub-1\,MB footprint.

3) We introduce a physics-aligned attention mechanism that maps temporal attention to the spectral domain and softly aligns it with analytical bearing fault-order bands. This provides auditable diagnostic evidence and improves robustness under structured interference and metadata uncertainty.

4) We couple the representation module with an EVT-calibrated decision layer and dual-threshold hysteresis. The resulting alarm policy maps streaming scores into stable maintenance triggers and supports target false-alarm intensity control under contaminated healthy calibration data.

5) We evaluate the framework with leakage-safe streaming protocols on three public benchmarks and an industrial pilot, reporting accuracy, timeliness, false-alarm intensity, transfer robustness, and edge latency under a unified deployment-oriented evaluation.

The remainder of this paper is organized as follows. Section~II presents the PG-TMT architecture, physics priors, and reliability-calibrated alarm layer. Section~III describes the datasets, streaming protocol, calibration setup, and evaluation metrics. Sections~IV and~V report online early warning results, robustness studies, cross-domain transfer, and ablations. Section~VI discusses industrial deployment and reproducibility, and Section~VII concludes the paper.

\section{Method: Physics-Guided Tiny-Mamba Transformer}
\label{sec:method}

\textsc{PG-TMT} is a compact encoder for \emph{online} early warning under class imbalance and domain shifts. It has three parts: a tri-branch sequence encoder for transient, long-range, and cross-channel cues; physics-guided priors that steer attention toward admissible fault-orders; and a reliability-calibrated decision layer that uses EVT and hysteresis to produce stable alarm episodes. For a $C$-channel window $\mathbf{x}_t\!\in\!\mathbb{R}^{C\times L}$ with hop $h\!\ll\!L$, the model outputs a calibrated anomaly score $s_t\!\in\![0,1]$, which is converted into online alarms with controlled false-alarm intensity.

\vspace{2pt}
\noindent\textbf{Why physics guidance matters.}
Physics guidance improves robustness to structured noise, supports transfer across speeds and sensors, makes attention spectra easier to interpret, and reduces score volatility for more stable EVT calibration.

\subsection{Architectural Overview}
\label{subsec:arch}

\begin{figure*}[t]
  \centering

  \subfloat[Architecture\label{fig:arch}]{%
    \includegraphics[width=0.59\textwidth]{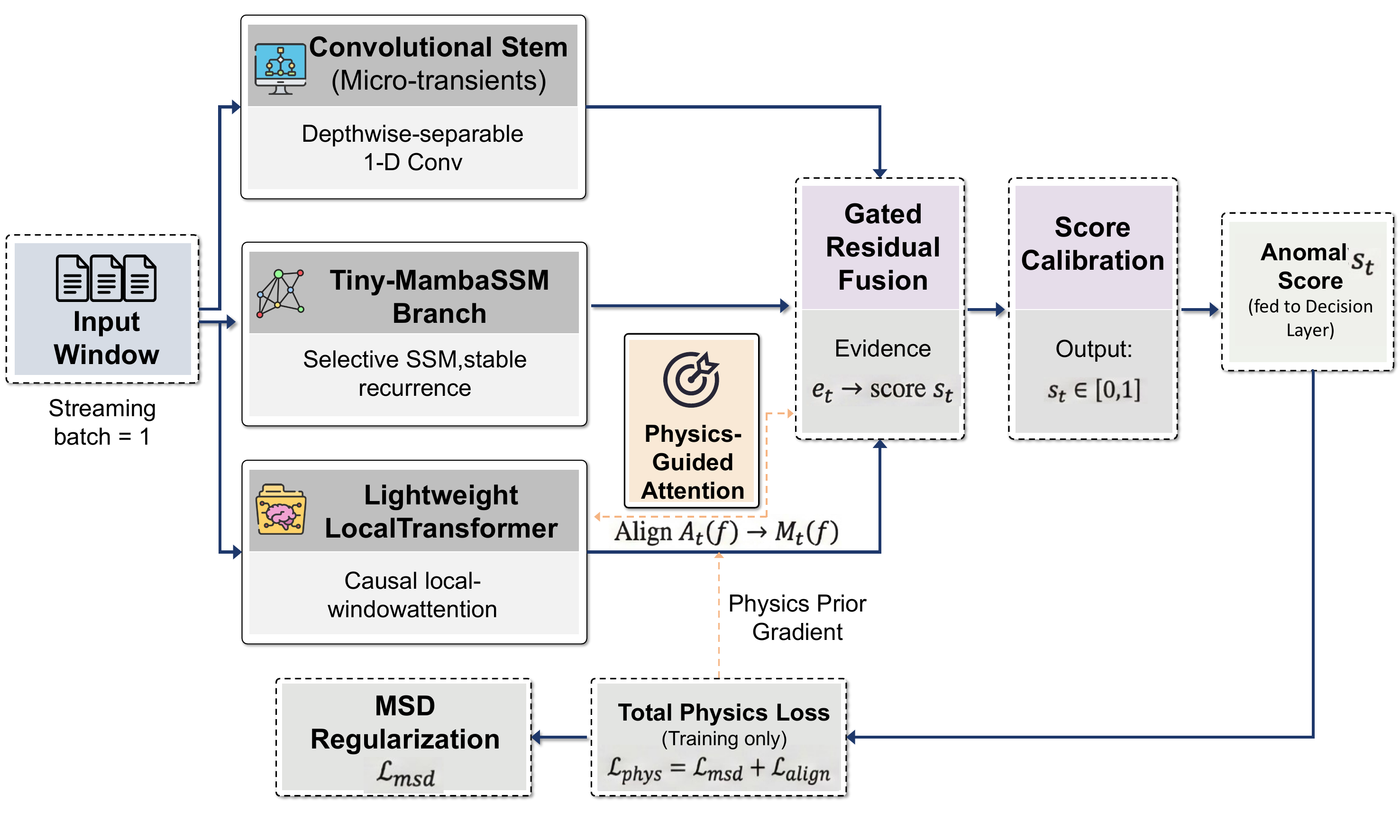}%
  }
  \hfill
  \subfloat[Reliability loop\label{fig:loop}]{%
    \includegraphics[width=0.41\textwidth]{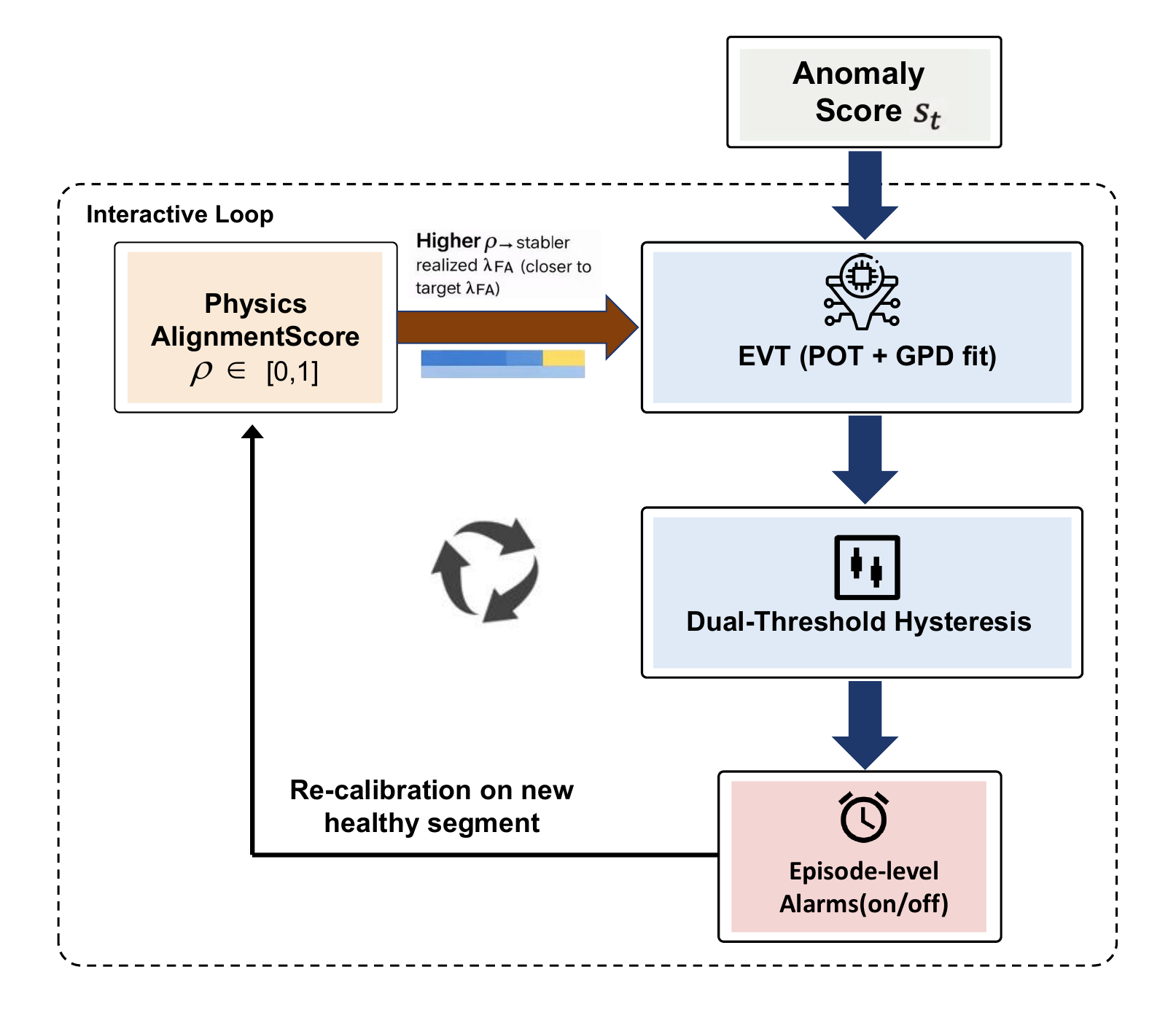}%
  }

  \caption{\textsc{PG-TMT} overview. (a) Architecture: a Tiny-Mamba SSM branch, a compact local-window Transformer, and a convolutional stem are fused into $\mathbf{r}_t$; attention association $p_t$ is tracked. (b) Reliability loop: physics alignment shapes $p_t$ (via a band mask), stabilizing the score tail that EVT models to calibrate $\tau_{\mathrm{on}}$; hysteresis converts scores into stable alarm episodes.}
  \label{fig:method_overview}
\end{figure*}

\textbf{Stem + two sequence branches.}
A depthwise-separable 1D convolutional stem extracts localized micro-transients; a Tiny-Mamba state-space branch models stable long-horizon dynamics using selective/structured SSMs~\cite{Gu2023Mamba,Dao2024SSD,Gu2022S4}; and a lightweight local-window Transformer captures cross-channel couplings~\cite{Vaswani2017Attention}. Their outputs are fused to produce association-discrepancy evidence $e_t$ and a calibrated score $s_t$ (Fig.~\ref{fig:method_overview}). The score is then mapped to streaming alarms by the EVT-based decision layer with hysteresis (Sec.~\ref{sec:decision}).

{1) Convolution stem (micro-transients).}
A short stack of depthwise-pointwise blocks (causal padding, optional dilation) preserves temporal resolution and outputs
$\mathbf{y}^{\mathrm{conv}}_t\!\in\!\mathbb{R}^{d_c}$, emphasizing impact-like transients and impulsive signatures.

{2) Tiny-Mamba SSM branch (long-horizon trends).}
We use selective/structured SSM layers to capture slow degradation with stable recurrence:
\begin{align}
\label{eq:ssm}
\mathbf{h}_{t+1} &= \mathbf{A}(\mathbf{g}_t)\,\mathbf{h}_t + \mathbf{B}(\mathbf{g}_t)\,\mathbf{u}_t,\\
\mathbf{y}^{\mathrm{ssm}}_{t} &= \mathbf{C}\,\mathbf{h}_t,
\end{align}
where $\mathbf{u}_t$ is a channel-reduced input, $\mathbf{h}_t\!\in\!\mathbb{R}^{d_{\text{ssm}}}$ is the latent state, and gates $\mathbf{g}_t$ modulate low-rank/diagonal $(\mathbf{A},\mathbf{B},\mathbf{C})$.

\emph{Stability parameterization.}
To ensure $\rho(\mathbf{A})<1$, we parameterize a continuous-time diagonal $\mathbf{A}_c=-\operatorname{softplus}(\boldsymbol{\eta})$ and use zero-order-hold discretization $\mathbf{A}=\exp(\Delta\,\mathbf{A}_c)$, $\mathbf{B}=\big(\int_{0}^{\Delta}\!\exp(\tau\mathbf{A}_c)\,d\tau\big)\mathbf{B}_c$, so $\Re(\lambda(\mathbf{A}_c))<0 \Rightarrow \rho(\mathbf{A})<1$. This yields bounded long-range memory with low-variance updates, important for stable tails in EVT calibration.

\emph{Note on nonlinearity (scope).}
The SSM branch is near-linear in its state update; strong nonlinear regimes (e.g., abrupt friction changes, severe impacts, rapidly evolving faults) can violate this assumption. In such cases, performance is supported by the convolution stem and attention branch, and we discuss this limitation and potential nonlinear SSM extensions in Sec.~\ref{sec:da-ablation}.

{3) Lightweight Transformer branch (cross-channel couplings).}
With causal neighborhood $\mathcal{N}_W(t)=\{\,i\mid t-W\le i\le t\,\}$ and $N_h$ heads, per-head attention is
\begin{align}
\label{eq:attn}
\alpha^{(h)}_{t,i} =
\frac{\exp\!\big(\langle \mathbf{q}^{(h)}_t,\mathbf{k}^{(h)}_i\rangle/\sqrt{d_{\text{head}}}\big)}
     {\sum_{j\in\mathcal{N}_W(t)} \exp\!\big(\langle \mathbf{q}^{(h)}_t,\mathbf{k}^{(h)}_j\rangle/\sqrt{d_{\text{head}}}\big)},
\quad i\in\mathcal{N}_W(t),
\end{align}
yielding $\mathbf{y}^{\mathrm{att}}_{t}\!\in\!\mathbb{R}^{d_a}$ that summarizes local cross-channel structure.

\textbf{4) Fusion and association-discrepancy evidence.}
We fuse $\mathbf{z}_t=[\mathbf{y}^{\mathrm{conv}}_t \| \mathbf{y}^{\mathrm{ssm}}_t \| \mathbf{y}^{\mathrm{att}}_t]$ using a gated residual:
\begin{align}
\label{eq:fuse}
\boldsymbol{\gamma}_t &= \sigma(\mathbf{W}_{\gamma}\mathbf{z}_t), \\
\mathbf{r}_t &= \boldsymbol{\gamma}_t\odot(\mathbf{W}_f\mathbf{z}_t)
               +(\mathbf{1}-\boldsymbol{\gamma}_t)\odot\mathbf{z}_t .
\end{align}
Aggregating \eqref{eq:attn} across heads yields a local association over time:
\begin{align}
\label{eq:pt}
p_t(i) = \tfrac{1}{N_h}\sum_{h=1}^{N_h}\alpha^{(h)}_{t,i},\qquad
\sum_{i\in\mathcal{N}_W(t)}p_t(i)=1 .
\end{align}
To ensure numerical stability, we smooth $p_t$ and a slow prior $\tilde p_t$ (EMA of past $p$) with the uniform $u_t$ on $\mathcal{N}_W(t)$: $\bar p_t=(1-\varepsilon)p_t+\varepsilon u_t$ and $\bar{\tilde p}_t=(1-\varepsilon)\tilde p_t+\varepsilon u_t$, with $\varepsilon\!\in\!(0,0.5)$.
We define the Jensen-Shannon discrepancy and evidence:
\begin{align}
\label{eq:disc}
\mathrm{disc}(p_t,\tilde p_t)
&= \tfrac{1}{2}\mathrm{KL}\!\big(\bar p_t \,\|\, m_t\big)
 + \tfrac{1}{2}\mathrm{KL}\!\big(\bar{\tilde p}_t \,\|\, m_t\big),\\
m_t&=\tfrac{1}{2}(\bar p_t+\bar{\tilde p}_t),\qquad 0\le \mathrm{disc}\le \log 2,\\
\label{eq:evidence}
e_t &= \mathbf{w}^{\top}\mathbf{r}_t+\lambda_{\mathrm{disc}}\cdot \mathrm{disc}(p_t,\tilde p_t),\\
\label{eq:score}
s_t &= \phi(e_t),\quad \phi(e)=\sigma(\kappa e+\beta),\ \kappa>0 .
\end{align}

\textbf{Calibration (post hoc).}
$\phi$ is a scalar, monotone calibrator acting on evidence $e_t$. We use temperature scaling~\cite{Guo2017Calibration} (learn $\kappa$ on held-out healthy/near-healthy data; set $\beta\!=\!0$) or isotonic regression under imbalance. This improves probability calibration while preserving ranking, and therefore preserves EVT tail fitting and the operating point.

\subsection{Physics-Guided Priors}
\label{subsec:priors}

We bias learning toward physically admissible manifolds while preserving data fit, so attention focuses on meaningful order harmonics and sidebands rather than spurious structured interference. Fig.~\ref{fig:band_alignment} visualizes physics-learning agreement; Table~\ref{tab:priors} summarizes priors and decision rules.

\subsubsection{MSD residual regularization}
\label{subsubsec:msd}

Assuming a lumped mass-spring-damper surrogate $m\ddot y+c\dot y+ky=f(t)+\varepsilon(t)$,
\begin{align}
\label{eq:msd}
r(t)&=m\ddot y+c\dot y+ky-f(t),\qquad
\mathcal{L}_{\mathrm{msd}}=\lambda_{\mathrm{msd}}\|r\|_2^2 .
\end{align}
\emph{Discretization.}
With sampling rate $F_s$, we use backward differences:
$\dot y\!\approx\!F_s\,(y_{n}-y_{n-1})$,
$\ddot y\!\approx\!F_s^2\,(y_{n}-2y_{n-1}+y_{n-2})$.
The nuisance driving term $f(t)$ is absorbed by the SSM readout, which learns latent driving-force dynamics from the stream, reducing cross-domain drift in the residual.
\emph{Linear surrogate.} The MSD term is used only as a lightweight first-order prior for dominant resonances and harmonic tracking. Unmodeled nonlinear effects are absorbed into $f(t)$ and handled by the Tiny-Mamba and Transformer branches, avoiding costly nonlinear solvers during online inference.

\subsubsection{Temporal attention to spectral attention (closed form)}
\label{subsubsec:temp2spec}

Let $F_s$ and grid $\{f_k\}_{k=0}^{N_f-1}\!\subset[0,F_s/2]$. We project $p_t$ to frequency via the normalized squared magnitude of a weighted DFT:
\begin{align}
\label{eq:spec-attn}
\tilde A_t(f_k) 
&=\left|\sum_{i\in\mathcal{N}_W(t)}
      p_t(i)\,e^{-j 2\pi f_k (i-i_0)/F_s}\right|^{2},\\[-2pt]
\intertext{where $i_0$ is the center index of $\mathcal{N}_W(t)$.}
A_t(f_k)
&= \frac{\tilde A_t(f_k)}{\sum_{k}\tilde A_t(f_k)} .
\end{align}

\indent This can be implemented via FFT with weights $p_t(i)$ at $O(W\log W)$ cost per step.

\subsubsection{Physics-derived band mask (soft, multi-band)}
\label{subsubsec:mask}

Given geometry $\{N_b,d,D_p,\theta\}$ and rotation $f_r$ (\emph{Hz} estimate $f_r=\hat{\Omega}/(2\pi)$; rpm $n=60\,f_r$), we define classical fault-orders:
\begin{align}
\label{eq:orders}
\mathrm{BPFI} &= \tfrac{N_b}{2}f_r\!\left(1+\tfrac{d}{D_p}\cos\theta\right),\\
\mathrm{BPFO} &= \tfrac{N_b}{2}f_r\!\left(1-\tfrac{d}{D_p}\cos\theta\right),\\
\mathrm{BSF}  &= \tfrac{D_p}{2d}f_r\!\left[1-\left(\tfrac{d}{D_p}\cos\theta\right)^{2}\right],\\
\mathrm{FTF}  &= \tfrac{1}{2}f_r\!\left(1-\tfrac{d}{D_p}\cos\theta\right).
\end{align}

\indent With sidebands $f_{j,m}=f_j+m f_r$ ($m=-K_s\ldots K_s$), the soft mask assigns larger weights to analytical fault-order bands by $M_t(f_\ell)\propto \sum_{j\in\mathcal{J}}\sum_{m=-K_s}^{K_s}w_{j,m}\exp\!\big(-(f_\ell-f_{j,m})^2/(2\sigma_j^2)\big)$, and is normalized so that $\sum_{\ell}M_t(f_\ell)=1$.

\textbf{\emph{Robustness to imperfect metadata.}}
The mask is intentionally soft (finite bandwidth $\sigma_j$) and multi-band (sum over $j \in \mathcal{J}$), so moderate errors in $\hat{\Omega}$ or geometry do not collapse the prior. We explicitly quantify this in Sec.~\ref{sec:results-online} via inference-time perturbations up to $\pm 15\%$ (Table~\ref{tab:sensitivity}), showing graceful degradation in PR--AUC rather than failure.

\emph{When speed sensing is unavailable.}
If no tachometer is available, $f_r$ can be estimated from vibration by dominant order tracking or envelope analysis~\cite{Antoni2007Cyclo,Yao2019TIM}. During uncertain intervals, we widen $\sigma_j$ and reduce the mask weight to avoid overconfident physics guidance.

\subsubsection{Spectral alignment loss (physics \texorpdfstring{$\rightarrow$}{->} attention)}
\label{subsubsec:align}

To avoid singularities, we smooth with uniform $u_f$: $\bar A_t=(1-\zeta)A_t+\zeta u_f$ and $\bar M_t=(1-\zeta)M_t+\zeta u_f$, with $\zeta\!\in\!(0,0.5)$.
We match distributions and penalize roughness:
\begin{align}
\label{eq:align}
\mathcal{L}_{\mathrm{align}}
&= \lambda_{\mathrm{align}}\,\mathrm{KL}\!\big(\bar M_t\,\big\|\,\bar A_t\big)
 + \lambda_{\mathrm{lap}}\sum_{k=0}^{N_{f}-1}\!\big|(\nabla_f \bar A_t)(f_k)\big|.
\end{align}
Since $A_t$ is the weighted-DFT image of $p_t$ in \eqref{eq:spec-attn}, minimizing \eqref{eq:align} shapes \emph{temporal} attention toward admissible fault-orders. This is the key pathway by which physics guidance influences the reliability loop: by confining attention to physically admissible bands, alignment reduces spurious spectral peaks (especially under structured noise) and keeps healthy-segment scores within a narrower, more stationary range, so the fitted generalized Pareto tail varies less across calibration windows. This lowers the variance of the EVT-derived threshold $\tau_{\mathrm{on}}$ in \eqref{eq:evt-thresh}, keeping the realized false-alarm intensity close to its target as operating conditions drift.
\noindent\textbf{Hyperparameter selection.}
We set the loss weights by a coarse grid search on the validation streams under the leakage-safe protocol (Sec.~\ref{subsec:validation}),
with the objective of maximizing PR--AUC at matched false-alarm intensity while avoiding unstable training.
Table~\ref{tab:loss-weights} reports the default values and the searched ranges used throughout the paper.

\begin{table}[t]
\centering
\caption{Loss-weight settings for physics-guided regularizers.}
\label{tab:loss-weights}
\begin{tabular}{lcc}
\hline
Weight & Default & Search range \\
\hline
$\lambda_{\mathrm{msd}}$ & 0.1 & $\{0,0.05,0.1,0.2,0.5\}$ \\
$\lambda_{\mathrm{align}}$ & 0.5 & $\{0,0.1,0.2,0.5,1.0\}$ \\
\hline
\end{tabular}
\end{table}

\noindent
Across datasets, performance is insensitive within the reported ranges, with $\lambda_{\mathrm{align}}$ mainly affecting alignment score stability and $\lambda_{\mathrm{msd}}$ primarily regularizing transient residuals; we use the default pair unless otherwise stated.

\subsubsection{Physics prior summary and alignment score}
\label{subsubsec:phys-summary}

The total physics prior is
\begin{align}
\label{eq:physics-total}
\mathcal{L}_{\mathrm{phys}}=\mathcal{L}_{\mathrm{msd}}+\mathcal{L}_{\mathrm{align}}.
\end{align}
We also track an overlap-based band-alignment score $\rho\in[0,1]$ between $A_t(f)$ and $M_t(f)$ (higher $\rho$ indicates attention concentrates within analytical order bands). In Sec.~\ref{sec:results-online}, $\rho$ decreases smoothly under metadata perturbations (Fig.~\ref{fig:param_sensitivity}), and removing physics guidance reduces $\rho$ while worsening robustness under structured noise (Table~\ref{tab:noise-structured}), providing quantitative evidence that alignment improves both interpretability and practical stability.

\begin{figure}[t]
  \centering
  \includegraphics[width=1\columnwidth]{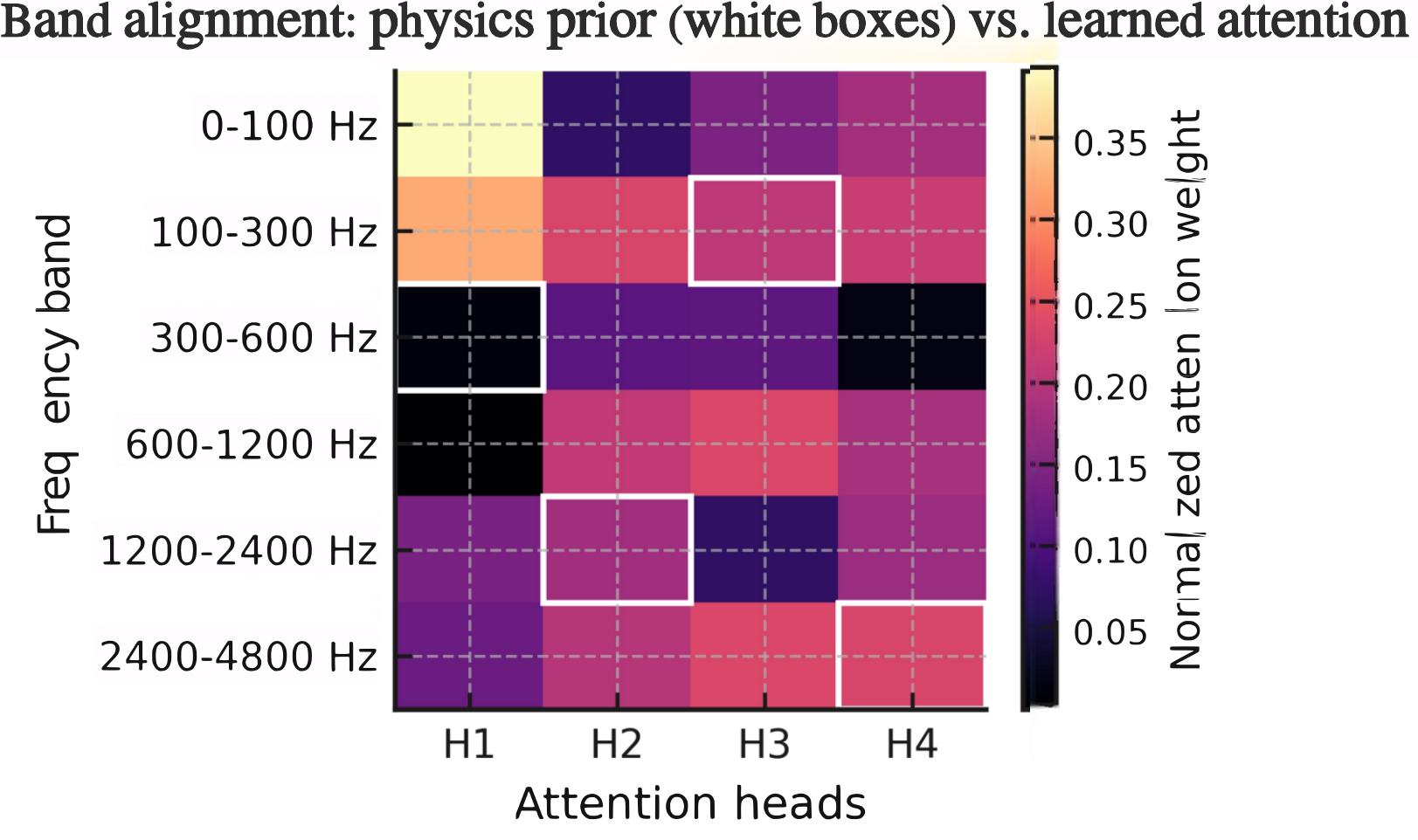}
  \caption{Physics-learning alignment. Heat shows agreement between $M_t(f)$ and the learned $A_t(f)$; bright bands near BPFI/BPFO/BSF/FTF (and sidebands) indicate physically meaningful focus.}
  \label{fig:band_alignment}
\end{figure}

\subsection{Decision Layer: EVT-Based Thresholding and Hysteresis}
\label{sec:decision}

\textbf{Notation.}
Let $\lambda_{\mathrm{FA}}$ denote the \emph{target false-alarm intensity} (episodes/hour), which is an operator-facing reliability requirement (acceptable nuisance alarm episodes per hour).
{We reserve} $\kappa$ for the calibrator slope in \eqref{eq:score}.

\subsubsection{Peaks-over-threshold modeling and target intensity}
\label{subsubsec:evt}

On healthy calibration segments, we model the \emph{tail} of $s_t$ by fitting a generalized Pareto distribution (GPD) to exceedances above a high level $u$ (POT setting)~\cite{Siffer2017SPOT,Vignotto2020GPD}. Let
\[
Y = s_t - u \;\big|\; s_t > u .
\]
Then for candidate threshold $\tau \ge u$,
\[
\Pr(s_t>\tau\mid s_t>u)
=
\bigl(1+\xi\,(\tau-u)/\beta\bigr)^{-1/\xi}_{+},
\]
and the exceedance intensity above $\tau$ is $\lambda_\tau = \lambda_u \Pr(s_t>\tau\mid s_t>u)$, where $\lambda_u$ is the exceedance arrival rate above $u$.

\textbf{Closed-form on-threshold.}
Solving $\lambda_\tau=\lambda_{\mathrm{FA}}$ gives
\begin{equation}
\label{eq:evt-thresh}
\tau_{\mathrm{on}}
= u + \frac{\beta}{\xi}\!\left[\left(\frac{\lambda_u}{\lambda_{\mathrm{FA}}}\right)^{\xi} - 1\right],
\qquad
\lim_{\xi\to 0}\tau_{\mathrm{on}}
= u + \beta \ln\!\left(\frac{\lambda_u}{\lambda_{\mathrm{FA}}}\right),
\end{equation}
with the usual validity domain $0<\lambda_{\mathrm{FA}}\le\lambda_u$ and $1+\xi(\tau_{\mathrm{on}}-u)/\beta>0$.

\textbf{Robust EVT calibration under imperfect ``healthy'' data.}
In practice, healthy calibration may be contaminated by incipient faults. To reduce sensitivity, we use a trimmed-tail fit: after selecting exceedances above $u$, we discard a small fraction of the most extreme exceedances before fitting $(\xi,\beta)$, which prevents a few contaminated, fault-like windows from dragging the tail fit and inflating $\tau_{\mathrm{on}}$. Sec.~\ref{sec:results-online} reports a dirty-calibration stress test (Table~\ref{tab:dirty-calibration}) showing stable thresholds and bounded FNR up to 20\% contamination.

\subsubsection{Dual thresholds, hold time, and alarm episodes (hysteresis)}
\label{subsubsec:hys}

We use dual-thresholds and a minimum hold time to suppress chatter:
\[
\mathrm{ALARM}_t=
\begin{cases}
1, & s_t \ge \tau_{\mathrm{on}},\\[2pt]
0, &
\begin{aligned}[t]
&s_t \le \tau_{\mathrm{off}}\coloneqq \tau_{\mathrm{on}}-\delta\\
&\text{and } t-t_{\text{on}}\ge T_{\min},
\end{aligned}\\[2pt]
\mathrm{ALARM}_{t-1}, & \text{otherwise.}
\end{cases}
\]
An \emph{alarm episode} is a maximal contiguous run of $\mathrm{ALARM}_t=1$. We also apply a short merging interval $\Delta T_{\mathrm{merge}}$ so adjacent episodes with smaller gaps are merged. The empirical false-alarm intensity $\hat{\lambda}_{\mathrm{FA}}$ is the number of alarm episodes per hour on healthy data, aligning the evaluation metric with operational requirements. The hold time $T_{\min}$ and merging interval together prevent a single noisy excursion from fragmenting into many short alarms, which would otherwise inflate the perceived alarm rate; counting at the episode level rather than per window therefore reflects the nuisance burden a maintenance team actually experiences.

\textbf{Condition-aware thresholds (rpm-aware).}
Because the score distribution and physical order locations vary with speed, a single static threshold can drift across operating conditions. We therefore allow weak dependence $\tau_{\mathrm{on}}(\hat{\Omega})$ via rpm binning or a slow piecewise log-linear mapping learned on healthy data, preserving the target $\lambda_{\mathrm{FA}}$ across speed regimes, while the physics mask updates the admissible bands through \eqref{eq:orders} and the soft mask definition. This is the second pathway by which physics guidance interacts with the reliability loop, stabilizing both score generation and the operating threshold under speed drift.

\subsection{Learning Objective and Calibration}
\label{subsec:objective}

\textbf{Task loss under imbalance.}
For binary labels $y_t\!\in\!\{0,1\}$, we use class-balanced BCE or focal loss to account for the rarity of fault windows.

\textbf{Total objective.}
\begin{align}
\label{eq:total-loss}
\mathcal{L}
&= \mathcal{L}_{\mathrm{task}} + \mathcal{L}_{\mathrm{phys}}.
\end{align}
When labels are unavailable (e.g., cold start), $\mathcal{L}_{\mathrm{task}}$ can be replaced by a one-class/self-supervised proxy (reconstruction or contrastive), consistent with healthy-only anomaly detection for industrial IoT~\cite{Liu2024IoTJAnoWGAN}, while keeping $\mathcal{L}_{\mathrm{phys}}$ active so physics constraints remain enforced.

\subsection{Complexity-Aware Design and Streaming Inference}
\label{subsec:complexity}

\textbf{Per-step computational cost.}
Depthwise convolution costs $O(C\,L\,k)$, the SSM update is $O(d_{\mathrm{ssm}})$, local-attention is $O(N_h\,W\,d_{\mathrm{head}})$, and the spectral alignment projection (FFT) costs $O(W\log W)$ with optional downsampling.

\textbf{Stateful streaming.}
We cache $(\mathbf{h}_t,\mathbf{K},\mathbf{V})$ and slide with hop $h\!\ll\!L$ to avoid redundant computation, tightening high-percentile latency tails and enabling stable real-time operation.

\textbf{Tooling.}
We export to ONNX/TensorRT with \texttt{batch}$=1$. FP16 kernels yield throughput gains with negligible changes in PR--AUC or MTTD.

\begin{table*}[!t]
\caption{Summary of physics-guided priors and the reliability decision layer}
\label{tab:priors}
\centering
\scriptsize
\setlength{\tabcolsep}{4pt}
\renewcommand{\arraystretch}{1.05}
\begin{tabular}{p{0.18\textwidth} p{0.39\textwidth} p{0.09\textwidth} p{0.28\textwidth}}
\toprule
\textbf{Component} & \textbf{Purpose} & \textbf{Eq./Def.} & \textbf{Key hyperparameters} \\
\midrule
MSD residual $\mathcal{L}_{\mathrm{msd}}$ &
Constrain physically plausible dynamics; reduce cross-domain drift; absorb unknown excitation via SSM state readout &
\eqref{eq:msd} &
$\lambda_{\mathrm{msd}}$; $(m,k,c)$; SSM readout settings \\
Spectral alignment $\mathcal{L}_{\mathrm{align}}$ &
Align learned spectral attention $A_t(f)$ with physics mask $M_t(f)$; suppress spurious structured peaks and stabilize tails &
\eqref{eq:spec-attn}, soft mask, \eqref{eq:align} &
$\lambda_{\mathrm{align}}$, $\lambda_{\mathrm{lap}}$, $N_f$, $K_s$, $\{\sigma_j,w_{j,m}\}$, $\hat{\Omega}(t)$ \\
EVT thresholding (GPD) &
Calibrate $\tau_{\mathrm{on}}$ to a target false-alarm intensity (episodes/hour) from healthy tails &
\eqref{eq:evt-thresh} &
$(u,\xi,\beta)$; $\lambda_u$; $\lambda_{\mathrm{FA}}$; trim ratio (robust fit) \\
Hysteresis + hold time &
Reduce chatter and fragmentation; define operational alarm episodes &
--- &
$\tau_{\mathrm{off}}=\tau_{\mathrm{on}}-\delta$; $T_{\min}$; $\Delta T_{\mathrm{merge}}$ \\
Task loss $\mathcal{L}_{\mathrm{task}}$ &
Learn decision-relevant score under imbalance &
--- &
$w_{+},w_{-}$ (or focal $\gamma$) \\
\bottomrule
\end{tabular}%
\end{table*}

\section{Datasets and Evaluation Protocols}
\label{sec:protocols}

We describe datasets, leakage-safe splits, streaming setup, noise, threshold calibration, and metrics for \emph{online} early-warning evaluation. The protocol is designed to: (i) prevent train--test leakage, (ii) reflect on-device streaming with \texttt{batch}=1, and (iii) report reliability-centered outcomes. Unless otherwise noted, \emph{all methods} use the same windowing, the same EVT+hysteresis decision stack, and the same evaluation code, so comparisons are made at \emph{matched} alarm intensity rather than heuristic thresholds. We follow standard PHM benchmarking practice and recent dataset audits for rotating machinery~\cite{Lei2016MSSPReview,Zhu2023MeasurementReview}.

\subsection{Datasets and Leakage Control}
\label{subsec:data}

To evaluate practical reliability under deployment-like constraints, we use three public rotating-machinery benchmarks and one industrial pilot (see Sec.~\ref{sec:industrial}). The pilot data are used for end-to-end validation only and are not used to tune model hyperparameters.

\textbf{CWRU (bearing faults).}
We include multiple loads, speeds, fault sizes, and sampling settings. Following prior benchmarking cautions, we construct split manifests so that each machine (or rig) and its operating-condition tuple (load, speed, sensor setup, sampling regime) appears in at most one of train/validation/test. This prevents condition leakage through repeated operating points~\cite{Hendriks2022CWRU,Smith2015CWRUStudy}.

\textbf{Paderborn (heterogeneous drive trains).}
We cover speed and torque grids across different drive trains and sensor configurations. Transfer tasks are built so that source and target differ in at least one factor (operating condition, sensor placement, or rig), reflecting realistic cross-domain deployment~\cite{Lessmeier2016Paderborn}.

\textbf{XJTU-SY (run-to-failure degradation).}
To verify early-warning reliability under progressive physical wear, we use complete degradation runs and enforce a chronological protocol: early healthy segments are used for representation learning and threshold calibration, while later life stages are held out for early-warning evaluation. Calibration and threshold fitting do not access future windows beyond the evaluation horizon~\cite{Lei2019XJTUSY_Tutorial}.

\noindent{Leakage control across all datasets}
\label{subsubsec:leakage}

We enforce a strict evaluation boundary aligned with deployment practice:
(i) splits are defined at the machine/rig level and, where applicable, further separated by operating conditions;
(ii) overlapping windows never cross split boundaries;
(iii) any resampling or filtering parameters are chosen within each split only;
(iv) channel normalization uses statistics computed on training windows only;
(v) EVT calibration uses only presumed healthy windows that temporally precede the evaluation horizon (Sec.~\ref{subsec:threshold});
(vi) all baselines and ablations reuse identical split manifests and preprocessing code to ensure fair comparisons under the same protocol.

{Practical note on calibration.}
Because calibration segments may contain incipient faults in real deployments, we include a contamination stress test and a robust EVT option with trimmed-tail fitting (Table~\ref{tab:dirty-calibration}). The protocol below specifies both standard and robust calibration pathways.

\subsection{Windows, Cold Start, and Streaming Setup}
\label{subsec:streaming}

Sliding windows of length $L$ and hop $h\!\ll\!L$ emulate online monitoring. A burn-in of $T_{\mathrm{burn}}$ windows initializes SSM states and attention KV caches and collects presumed healthy statistics for calibration; decisions during burn-in are ignored. After burn-in, streaming inference runs at \texttt{batch}=1 with state caching and fixed hop.

\begin{figure}[!t]
  \centering
  \includegraphics[width=1.0\columnwidth]{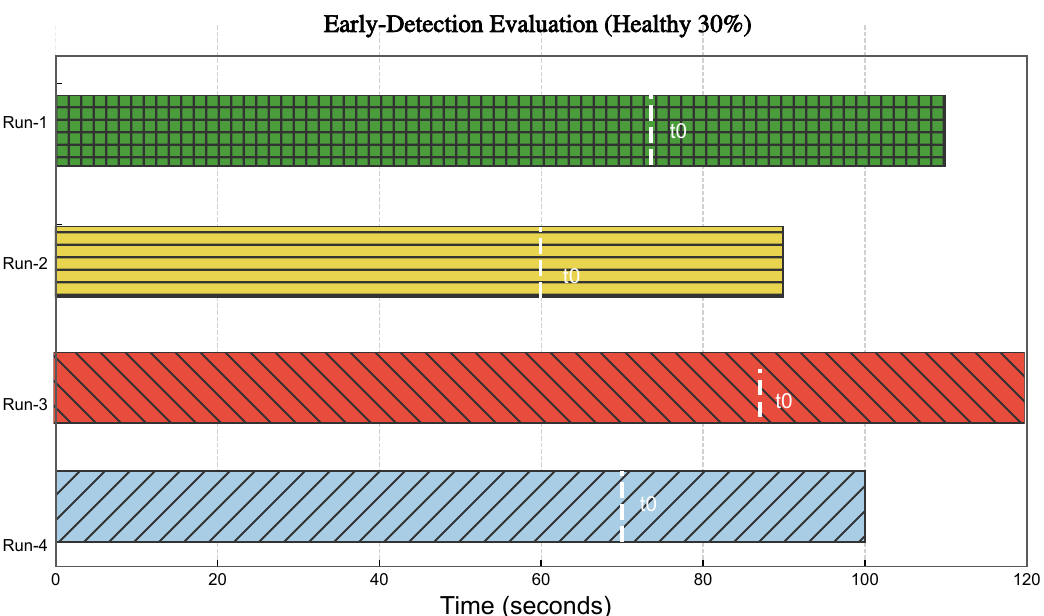}
  \caption{Streaming timeline. $t_{\mathrm{phys}}$: first physically detectable deviation (label or expert annotation); $t_0$: first issued alarm under hysteresis (Sec.~\ref{subsec:threshold}). Windows contributing to PR--AUC/MTTD/FAR are highlighted. A merging interval $\Delta T_{\mathrm{merge}}$ merges nearby onsets; runs with no alarm by horizon end are right-censored.}
  \label{fig:timeline}
\end{figure}

\noindent{Reference times and censoring.}
\label{subsubsec:ref-times}
For run $r$, let $t_{\mathrm{phys}}^{(r)}$ denote the first \emph{physically detectable} deviation time and
$t_{0}^{(r)}$ the first alarm time under the dual-threshold policy.
If no alarm is triggered before the evaluation horizon $T$, the run is treated as right-censored.
We summarize the lead time $\Delta^{(r)} \!=\! t_{0}^{(r)} - t_{\mathrm{phys}}^{(r)}$ with censoring-aware estimators
(Sec.~\ref{subsec:metrics}).
\subsection{Domain-Shift Tasks (Cross-Load/Speed/Sensor/Machine)}
\label{subsec:shift}

We evaluate three shift families:
(i) cross-load/speed within a dataset,
(ii) cross-sensor/rig,
(iii) cross-dataset (e.g., CWRU$\to$Paderborn, Paderborn$\to$XJTU-SY).
Each directed task reports source-only, post-adaptation, and retention (accuracy and timeliness), following transfer-learning and domain adaptation practice~\cite{Ganin2016JMLR}. To connect with recent domain-generalization practice in rotating machinery, we also report transfer under matched alarm intensity and discuss robustness trends under speed/load/sensor variability~\cite{Xiao2025DGSurvey,Jia2024CDDG,Zhao2024IoTJDyEdgeGAT}.

\textbf{Metadata uncertainty (speed/geometry).}
Real deployments often have imperfect rotational speed and bearing geometry. Rather than assuming perfect metadata, we explicitly evaluate inference-time perturbations of $\hat{\Omega}$ and geometry by up to $\pm 15\%$(Table~\ref{tab:sensitivity}) to quantify graceful degradation under mask mismatch.

\subsection{Noise Model and SNR Setting}
\label{subsec:noise}

\subsubsection{Additive noise (SNR sweeps)}
\label{subsubsec:snr}

To probe robustness, we add zero-mean white noise to reach $\mathrm{SNR}\in\{0,5,10,15,20\}\,\mathrm{dB}$. For $\mathbf{x}\!\in\!\mathbb{R}^{C\times L}$ with per-sample, per-channel average power $\|\mathbf{x}\|_F^2/(CL)$,
\[
\mathrm{SNR}_{\mathrm{dB}}=10\log_{10}\!\left(\frac{\|\mathbf{x}\|_F^2/(CL)}{\sigma_n^2}\right),
\]
so we set $\sigma_n^2=\big(\|\mathbf{x}\|_F^2/(CL)\big)\cdot 10^{-\mathrm{SNR}_{\mathrm{dB}}/10}$ per window. Noise is injected after training-set normalization; unless stated, channels receive independent draws.

\subsubsection{Structured industrial noise (0\,dB)}
\label{subsubsec:structured-noise}

Industrial environments frequently include structured interference that is not well modeled by white noise. We therefore include a structured-noise stress test at 0\,dB SNR (Table~\ref{tab:noise-structured}) with:
(i) pink noise (1/f),
(ii) 50\,Hz power-line interference,
(iii) low-frequency drift.
Each perturbation is scaled to match the target SNR using the same power definition as above. This protocol is designed to expose failure modes where data-driven models confuse periodic interference (e.g., 50\,Hz) with fault harmonics.

\subsection{Threshold Calibration and Hysteresis}
\label{subsec:threshold}

Thresholds are calibrated on presumed healthy (or near-healthy) windows via EVT under a peaks-over-threshold (POT) model~\cite{Siffer2017SPOT,Vignotto2020GPD}. Let $u$ be a high preliminary level; exceedances
$Y=s_t-u\mid s_t>u$
follow a $\mathrm{GPD}(\xi,\beta)$, and exceedance arrivals above $u$ are modeled as a Poisson process with rate $\lambda_u$ (episodes/hour). For a target false-alarm intensity $\lambda_{\mathrm{FA}}$ (episodes/hour), the on-threshold is
\[
\tau_{\mathrm{on}}
= u+\frac{\beta}{\xi}\!\left[\left(\frac{\lambda_u}{\lambda_{\mathrm{FA}}}\right)^{\xi}-1\right],
\qquad
\lim_{\xi\to 0}\ \tau_{\mathrm{on}} = u+\beta\ln\!\frac{\lambda_u}{\lambda_{\mathrm{FA}}}.
\]

This is valid when $0<\lambda_{\mathrm{FA}}\le \lambda_u$ and $1+\xi(\tau_{\mathrm{on}}-u)/\beta>0$. We apply $\tau_{\mathrm{off}}=\tau_{\mathrm{on}}-\delta$ with $\delta>0$ and a minimum hold time $T_{\min}$ to implement hysteresis.

\textbf{Robust calibration under contaminated ``healthy'' data.}
To handle the case where calibration data contain incipient faults, we additionally evaluate a trimmed-tail EVT fit: after selecting exceedances above $u$, we discard a small fraction of the most extreme exceedances before fitting $(\xi,\beta)$. This reduces sensitivity to a small number of fault-like windows while preserving the tail shape under normal operation. We report a contamination stress test in Table~\ref{tab:dirty-calibration}.

\textbf{Declustering and episode counting.}
We apply a merging interval $\Delta T_{\mathrm{merge}}$ to merge nearby alarms into a single episode. This aligns evaluation with operator-facing nuisance episodes and reduces dependence violations by declustering extreme excursions, improving the practical fit of the Poisson-arrival assumption in EVT.

\textbf{Condition-aware calibration.}
When speed drifts, we optionally maintain $\tau_{\mathrm{on}}(\hat{\Omega})$ across rpm bins using healthy data to stabilize the realized $\widehat{\lambda}_{\mathrm{FA}}$.

\subsection{Additional Stress Test: Proxy Compound Faults}
\label{subsec:compound}

Real machines may exhibit compound faults (e.g., inner + outer race). When datasets lack naturally labeled compound cases, we construct a proxy stress test by superposing synchronized windows from two single-fault conditions (IR and OR) after power normalization:
\[
\mathbf{x}_{\mathrm{IR+OR}} = \alpha\,\mathbf{x}_{\mathrm{IR}} + (1-\alpha)\,\mathbf{x}_{\mathrm{OR}},
\]
with $\alpha$ balancing component energies and optional noise injected after mixing. We report PR--AUC, ROC--AUC, and MTTD at matched alarm intensity, and the relative drop versus the single-fault baseline (Table~\ref{tab:compound-ablation}). This exposes feature confusion under overlapping fault harmonics and tests whether physics-guided masking preserves multi-band focus.

\subsection{Primary Metrics}
\label{subsec:metrics}

\textbf{PR--AUC and ROC--AUC.}
We report both, emphasizing PR--AUC under class imbalance~\cite{Saito2015PRvsROC}. Scores $s_t$ are evaluated over streaming windows; operating points for timing metrics are taken at matched alarm intensity.

\textbf{MTTD (lead time).}
For run $r$, $\Delta^{(r)}=t_0^{(r)}-t_{\mathrm{phys}}^{(r)}$. Because misses induce right censoring, we summarize $\{\Delta^{(r)}\}$ via Kaplan--Meier curves and report mean/median lead time with confidence intervals; group comparisons may use Cox proportional-hazards models (hazard ratios)~\cite{Kaplan1958JASA,Cox1972JRSSB}.

\textbf{False-alarm intensity (FAR).}
Episodes are formed by hysteresis and then merged with $\Delta T_{\mathrm{merge}}$; FAR is the number of unique episodes per hour on healthy data. We denote the target intensity by $\lambda_{\mathrm{FA}}$ and its empirical estimate by $\widehat{\lambda}_{\mathrm{FA}}$.

\textbf{Latency.}
Per-window inference with \texttt{batch}=1 after warm-up; we report p50/p90/p99 and sustainable FPS on CPU/Jetson. Latencies include preprocessing, model forward, and EVT+hysteresis evaluation.

\textbf{Calibration and significance.}
Reliability diagrams and expected calibration error quantify probability calibration~\cite{Guo2017Calibration}. AUC comparisons use a standard correlated-ROC significance test, and multiple comparisons are controlled with false-discovery-rate correction. Where relevant, Numenta-style anomaly scoring complements timeliness and nuisance-cost reporting.

\subsection{Validation Protocol and Confidence Intervals}
\label{subsec:validation}

Hyperparameters are tuned on a machine/rig-level validation split to prevent leakage. Each configuration is repeated with multiple seeds; unless noted, all methods share the same windowing, calibration, and evaluation code. We report mean $\pm$ 95\% confidence intervals over runs/seeds. PR--AUC and MTTD intervals are bootstrapped at the run level. For AUC comparisons, $p$-values come from a correlated-ROC significance test with FDR correction; for timing, we complement summary statistics with KM curves and Cox hazard ratios.

\subsection{Reproducibility Artifacts}
\label{subsec:repro}

We release preprocessing scripts, split manifests, and configuration files for online evaluation (including EVT, hysteresis, and merging parameters), together with ONNX/TensorRT exporters and latency/FPS harnesses. We provide commit hashes, seed files, and environment manifests to support exact replication of figures and tables in the Results and Ablations.

\section{Online Early-Warning Results}
\label{sec:results-online}

We evaluate PG-TMT under the leakage-safe streaming protocol in Section~\ref{sec:protocols} with \texttt{batch}$=1$, reflecting realistic embedded deployment. Beyond predictive accuracy, we stress four properties that determine whether early warning is \emph{operationally reliable}: (i) real-time feasibility, (ii) robustness to industrial perturbations, (iii) tolerance to metadata uncertainty, and (iv) statistical stability of threshold calibration.

All operating thresholds are calibrated using EVT with hysteresis at \emph{matched} false-alarm intensity (episodes/hour)~\cite{Siffer2017SPOT,Vignotto2020GPD}. Metrics follow imbalance-aware best practice~\cite{Saito2015PRvsROC}. Our robustness tests target both additive noise and structured interference that commonly exhibits cyclostationary or narrowband characteristics in rotating machinery signals~\cite{Antoni2007Cyclo,Randall2011Tutorial}.

\subsection{Latency and Streaming Feasibility}
\label{subsec:latency}

\begin{figure}[!t]
  \centering
  \includegraphics[width=\columnwidth]{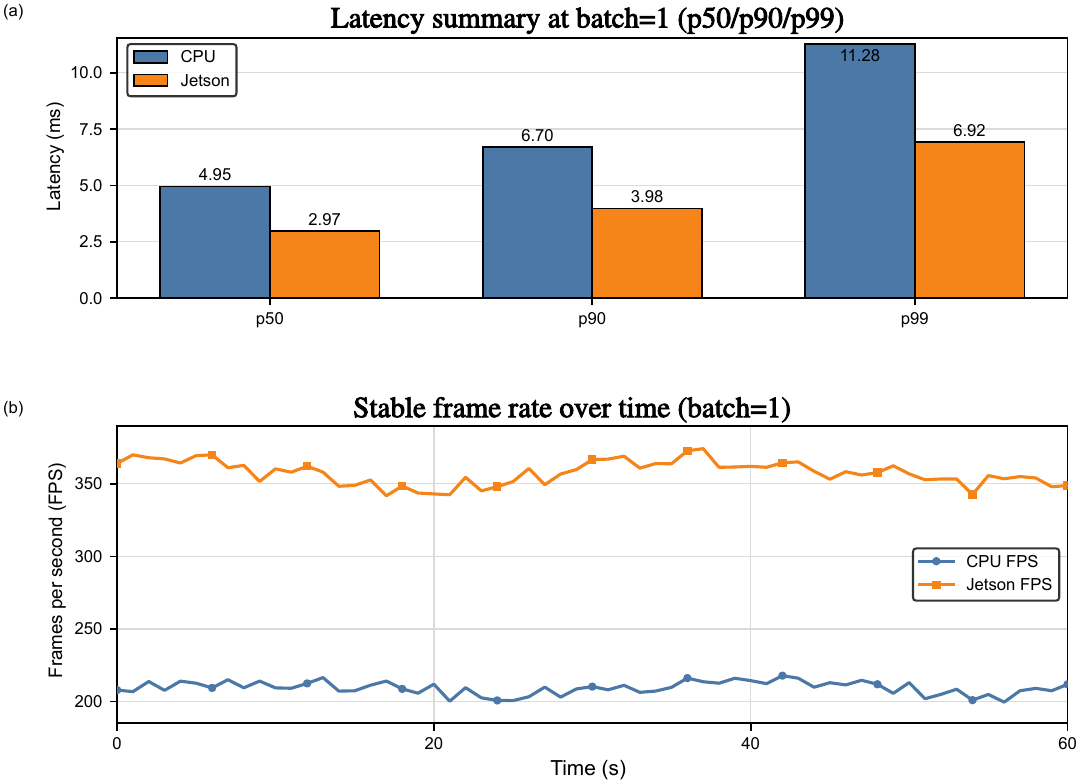}
  \caption{Deployment metrics at \texttt{batch}=1 on CPU and Jetson. (a) Latency summary (p50/p90/p99). (b) Stable frame rate over time (FPS).}
  \label{fig:deploy_metrics}
\end{figure}

Figure~\ref{fig:deploy_metrics} summarizes runtime at \texttt{batch}=1 on a desktop CPU and an embedded Jetson device. PG-TMT exhibits tight latency distributions with narrow high-percentile tails (p90/p99), important for stable and predictable streaming decisions. This is enabled by stateful Tiny-Mamba updates and local-attention with KV caching, consistent with efficient SSM formulations~\cite{Gu2023Mamba,Dao2024SSD}. The sustained frame rate stays well above dataset acquisition rates, leaving headroom for online EVT, hysteresis, and logging (Table~\ref{tab:latency-complexity}), confirming that reliability-calibrated early warning is feasible under strict \texttt{batch}=1 streaming without sacrificing interpretability or decision control.

\subsection{Robustness to Additive and Structured Noise}
\label{subsec:noise}

\subsubsection*{Additive noise (SNR sweeps)}
\begin{figure}[!t]
  \centering
  \includegraphics[width=\columnwidth]{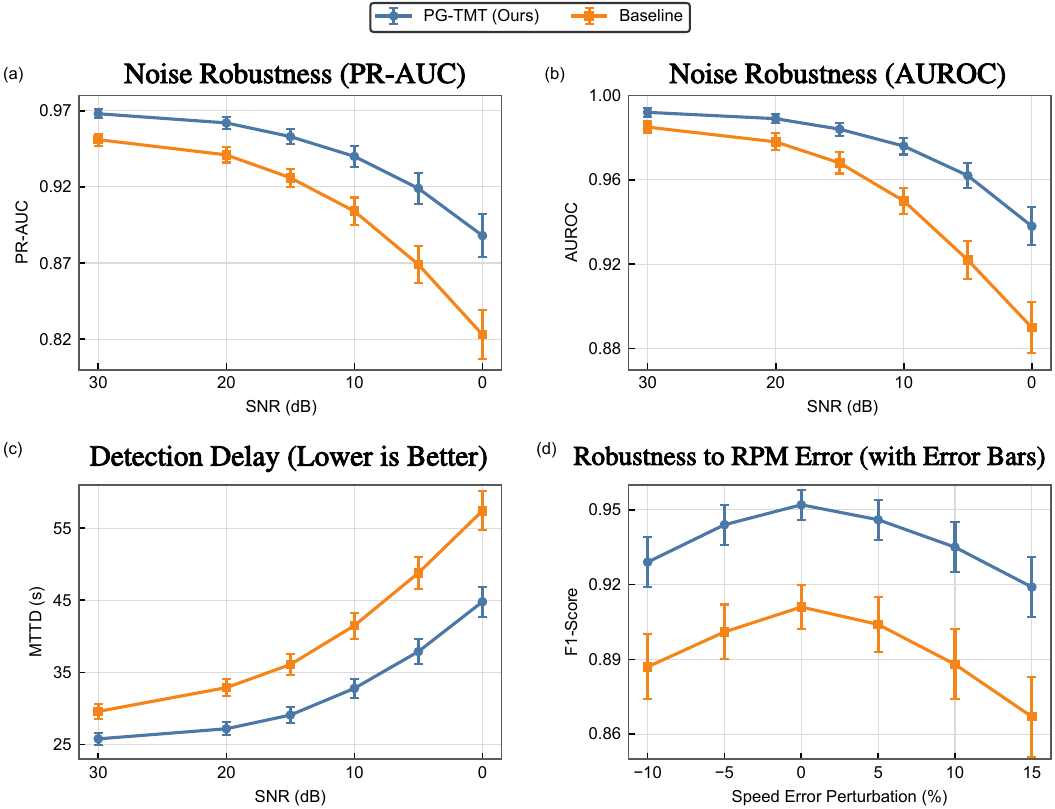}
    \caption{Robustness stress tests. (a) PR--AUC and (b) AUROC under additive noise across SNR. (c) Detection delay (MTTD) under SNR sweeps (lower is better). (d) Sensitivity to speed estimation error with error bars.}
  \label{fig:robustness}
  
\end{figure}

We evaluate robustness under SNR sweeps by calibrating thresholds at a high-SNR reference (e.g., 20\,dB) and keeping them fixed across noise levels. As SNR decreases, PG-TMT degrades gracefully rather than collapsing (Fig.~\ref{fig:robustness}): PR--AUC and AUROC decline smoothly, and detection delay (MTTD) increases in a controlled manner. This behavior reflects robustness from both the representation layer (tri-branch encoding) and the decision stack (EVT-based calibration with hysteresis), which stabilizes streaming alarms under noisy operating regimes~\cite{Siffer2017SPOT,Vignotto2020GPD}.

\subsubsection*{Structured industrial noise (0\,dB)}
Industrial environments often contain structured disturbances such as power-line interference, slow drift, and $1/f$ noise. These effects are not well represented by AWGN and can induce spurious periodic signatures that contaminate vibration-based diagnostics~\cite{Antoni2007Cyclo,Randall2011Tutorial}. We therefore evaluate structured disturbances at 0\,dB SNR, including pink noise ($1/f$), 50\,Hz power-line interference, and low-frequency drift. Results are summarized in Table~\ref{tab:noise-structured}.

\begin{table}[t]
\centering
\caption{Robustness under structured noise (0\,dB SNR). PR--AUC comparison.}
\label{tab:noise-structured}
\small
\setlength{\tabcolsep}{4pt}
\begin{tabular}{lccc}
\hline
Noise type & \textbf{PG-TMT (Ours)} & w/o physics & Std.\ Mamba \\
\hline
Gaussian (AWGN)     & \textbf{0.862} & 0.810 & 0.795 \\
Pink noise ($1/f$)  & \textbf{0.885} & 0.740 & 0.715 \\
Power line (50\,Hz) & \textbf{0.912} & 0.655 & 0.630 \\
Low-freq drift      & \textbf{0.934} & 0.825 & 0.810 \\
\hline
\end{tabular}
\end{table}

The advantage widens substantially under structured interference. Under 50\,Hz contamination, PG-TMT maintains PR--AUC above 0.9, while non-physics baselines drop to around 0.63--0.66. This reflects a practical failure mode: without physics-guided masking, persistent narrowband interference can be misinterpreted as fault-related periodicity~\cite{Randall2011Tutorial,Yao2019TIM}. In contrast, the order-band mask suppresses structured harmonic energy before it propagates into the attention association and anomaly score, leading to materially stronger robustness under factory-like conditions.

\subsection{\texorpdfstring{Sensitivity to Speed and Geometry Uncertainty}{Sensitivity to Speed and Geometry Uncertainty}}
\label{subsec:sensitivity}

\begin{figure}[t]
  \centering
  \includegraphics[width=\columnwidth]{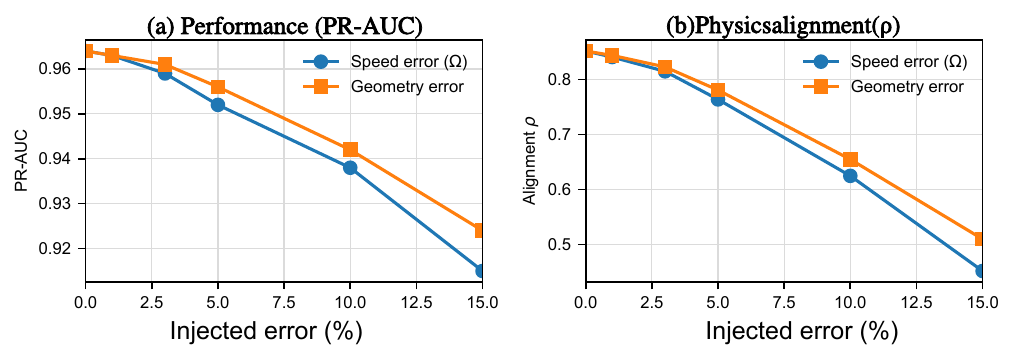}
  \caption{Sensitivity to metadata uncertainty. PR--AUC (left) and physics alignment score $\rho$ (right) under inference-time perturbations of rotational speed $\Omega$ and bearing geometry used by the order-band mask. Performance degrades smoothly up to $\pm 15\%$ error, while $\rho$ decreases as expected due to band mismatch.}
  \label{fig:param_sensitivity}
\end{figure}

In real deployments, rotational speed and bearing geometry are often imperfectly specified. Since classical bearing-order features depend on both, tolerance to such metadata uncertainty matters in practice~\cite{Randall2011Tutorial,McFadden1984JSV}. We inject inference-time perturbations up to $\pm 15\%$ into the speed and geometry used to build the order-band mask. Figure~\ref{fig:param_sensitivity} summarizes trends in performance and physics alignment; Table~\ref{tab:sensitivity} reports the values.

\begin{table}[t]
\centering
\caption{Sensitivity to speed and geometry errors (inference-time perturbation).}
\label{tab:sensitivity}
\small
\setlength{\tabcolsep}{4pt}
\begin{tabular}{c|cc|cc}
\hline
\multirow{2}{*}{Error (\%)} 
& \multicolumn{2}{c|}{Speed error ($\Omega$)} 
& \multicolumn{2}{c}{Geometry error} \\
 & PR--AUC & $\rho$ & PR--AUC & $\rho$ \\
\hline
0  & 0.964 & 0.852 & 0.964 & 0.852 \\
5  & 0.952 & 0.764 & 0.956 & 0.781 \\
10 & 0.938 & 0.625 & 0.942 & 0.655 \\
15 & 0.915 & 0.452 & 0.924 & 0.510 \\
\hline
\end{tabular}
\end{table}
Even at $\pm 15\%$ metadata error, PR--AUC remains above 0.91. As expected, the alignment score $\rho$ decreases when analytical bands are perturbed, because the mask becomes less matched to the true fault-order structure. Importantly, predictive performance degrades smoothly instead of collapsing. We attribute this to two design choices. First, the mask is soft and uses finite bandwidth, so modest band mismatch does not eliminate informative energy. Second, the convolutional stem and the SSM branch provide complementary cues that do not rely entirely on precise order-band placement. Overall, PG-TMT does not require perfectly known physical parameters and remains effective under realistic speed and geometry uncertainty.

\subsection{Alarm Stability and Robust Threshold Calibration}
\label{subsec:evt-robust}

EVT thresholds are calibrated on presumed healthy windows. In practice, incipient faults may already be present during calibration, contaminating the tail fit and destabilizing the operating point. We therefore simulate contaminated calibration by injecting up to 20\% incipient-fault samples into the healthy set.
Table~\ref{tab:dirty-calibration} reports threshold drift and the resulting false-negative rate (FNR), extending standard POT-based stream calibration to a contamination-robust setting~\cite{Siffer2017SPOT,Vignotto2020GPD}.

\begin{table}[t]
\centering
\caption{Robustness of EVT Threshold Calibration under Contaminated ``Healthy'' Data.}
\label{tab:dirty-calibration}
\small
\setlength{\tabcolsep}{4pt}
\begin{tabular}{c|cc|cc}
\hline
\multirow{2}{*}{Contam.\ (\%)} 
& \multicolumn{2}{c|}{Standard EVT} 
& \multicolumn{2}{c}{\textbf{Robust EVT (Ours)}} \\
 & $\tau$ & FNR (\%) & $\tau$ & FNR (\%) \\
\hline
0  & 0.750 & 1.5  & 0.750 & \textbf{1.5} \\
5  & 0.785 & 2.8  & 0.758 & \textbf{1.7} \\
10 & 0.852 & 6.5  & 0.765 & \textbf{2.1} \\
20 & 0.945 & 14.2 & 0.788 & \textbf{3.8} \\
\hline
\end{tabular}
\end{table}

Standard EVT exhibits pronounced threshold drift and an FNR explosion under contamination. In contrast, trimmed-tail calibration maintains stable thresholds and bounded FNR even at 20\% contamination. This directly supports the reliability claim: the decision layer remains statistically stable under imperfect health labeling, which is common in real condition monitoring pipelines.

\subsection{Proxy Compound Fault Stress Test}
\label{subsec:compound}

{Compound faults can create overlapping spectral signatures and confound purely data-driven models.}
{To probe this challenge, we conduct a proxy compound-fault stress test by constructing IR+OR mixtures.}
{The key question is whether the model remains stable when multiple fault signatures overlap in the spectrum, a known difficulty in bearing diagnostics}~\cite{Randall2011Tutorial,McFadden1984JSV}.
Table~\ref{tab:compound-ablation} summarizes the aggregated results.

\begin{table}[t]
\centering
\caption{Stress test and ablation summary.}
\label{tab:compound-ablation}
\footnotesize
\setlength{\tabcolsep}{3pt}
\renewcommand{\arraystretch}{1.12}
\begin{tabular}{p{0.29\columnwidth} p{0.48\columnwidth} p{0.18\columnwidth}}
\toprule
\textbf{Variant} & \textbf{Primary effect} & \textbf{Metric trend} \\
\midrule
Compound IR+OR &
Overlapping inner-race and outer-race signatures create feature confusion. PG-TMT reduces the PR drop from 14.1\% to 3.7\%. &
PR$\downarrow$, MTTD$\uparrow$ \\

No physics priors &
Attention leaks beyond BPFI/BPFO/BSF/FTF bands and becomes more sensitive to structured interference. &
PR$\downarrow$, FAR$\uparrow$, $\rho\downarrow$ \\

No Mamba branch &
Long-horizon degradation tracking becomes weaker. &
MTTD$\uparrow$, PR$\downarrow$ \\

No Transformer &
Cross-channel resonance coupling is under-modeled. &
PR$\downarrow$, ROC$\downarrow$ \\

No conv stem &
Impact-like transient sensitivity is reduced. &
MTTD$\uparrow$ \\

No EVT &
A fixed threshold cannot match false-alarm intensity across domains. &
FAR$\uparrow$ \\

No hysteresis &
Short score flickers are not merged into stable alarm episodes. &
FAR$\uparrow$ \\
\bottomrule
\end{tabular}
\end{table}

In the proxy compound-fault stress test, the baseline without physics guidance suffers a 14.1\% PR drop, whereas PG-TMT limits the drop to 3.7\% and keeps MTTD close to the single-fault setting. This suggests that physics-guided attention helps preserve multi-band fault evidence when inner-race and outer-race signatures overlap.

Table~\ref{tab:compound-ablation} summarizes the main ablation findings. Removing the physics priors causes the largest degradation in PR and alignment score, confirming that order-band guidance is central to robustness and interpretability. Removing the Mamba branch mainly increases MTTD, indicating weaker long-horizon degradation tracking. Removing the Transformer branch or convolutional stem degrades accuracy and timeliness. The decision-layer ablations reveal a different failure mode: a fixed threshold preserves ranking metrics but increases FAR, while removing hysteresis causes alarm fragmentation.

\subsection{Qualitative Early-Warning Behavior}
\label{subsec:qual}

On run-to-failure (XJTU-SY) and staged-fault sequences, PG-TMT triggers earlier and more stable alarm episodes relative to the physical onset $t_{\mathrm{phys}}$ (Fig.~\ref{fig:timeline}). As degradation progresses, attention concentrates near defect orders, the anomaly score rises coherently, and EVT with hysteresis converts evolving patterns into actionable alarms while suppressing fragmentation.

\medskip
\noindent\textbf{Summary.}

Under strict streaming constraints, PG-TMT demonstrates predictable real-time latency, strong robustness to structured industrial interference, graceful degradation under speed and geometry uncertainty, stable EVT calibration under contaminated health data, and improved robustness under proxy compound faults. Together, these results support the practical feasibility of PG-TMT for deployment-oriented early warning under the evaluated conditions.

\section{Cross-Domain/Sensor Adaptation and Ablations}
\label{sec:da-ablation}

We assess transfer robustness under the leakage-safe protocol in Section~\ref{sec:protocols} and isolate key PG-TMT design factors via ablations. We include standard transfer learning and domain adaptation baselines~\cite{Ganin2016JMLR} and strong supervised references from rotating machinery fault diagnosis~\cite{Shao2019TII_DTL,Xiao2024BVTransformer}. We also consider recent domain generalization lines that target nonstationary machinery and IIoT environments~\cite{Jia2024CDDG,Xiao2025DGSurvey,Zhao2024IoTJDyEdgeGAT}.

\subsection{Baselines and Recent Related Models}
We also compare against (or discuss, when reproduction is infeasible) recent rotating-machinery models built on Mamba or SSM-style backbones and Transformer-Mamba hybrids~\cite{Yi2025VibrMamba,Zhang2025GsXANet,Xia2025BMTMNet}, and OOD-oriented augmentation under zero-faulty data~\cite{Chen2025SMRN}. All methods share identical preprocessing and the same post-hoc decision pipeline (windowing, EVT calibration, hysteresis, episode merging), so evaluation at matched false-alarm intensity avoids threshold-induced confounding and makes retention and gain more indicative of representation robustness.

\subsection{Matched-Intensity Decision Control}
On each target domain, $\tau_{\mathrm{on}}$ is re-estimated on target healthy or near-healthy windows to enforce the same $\lambda_{\mathrm{FA}}$, since a threshold calibrated once can drift across operating regimes with speed, load, or sensor changes. This matched-intensity protocol ensures retention and gain mainly reflect representation robustness rather than threshold mismatch~\cite{Siffer2017SPOT,Vignotto2020GPD}.

\begin{figure}[!t]
  \centering\includegraphics[width=0.5\textwidth]{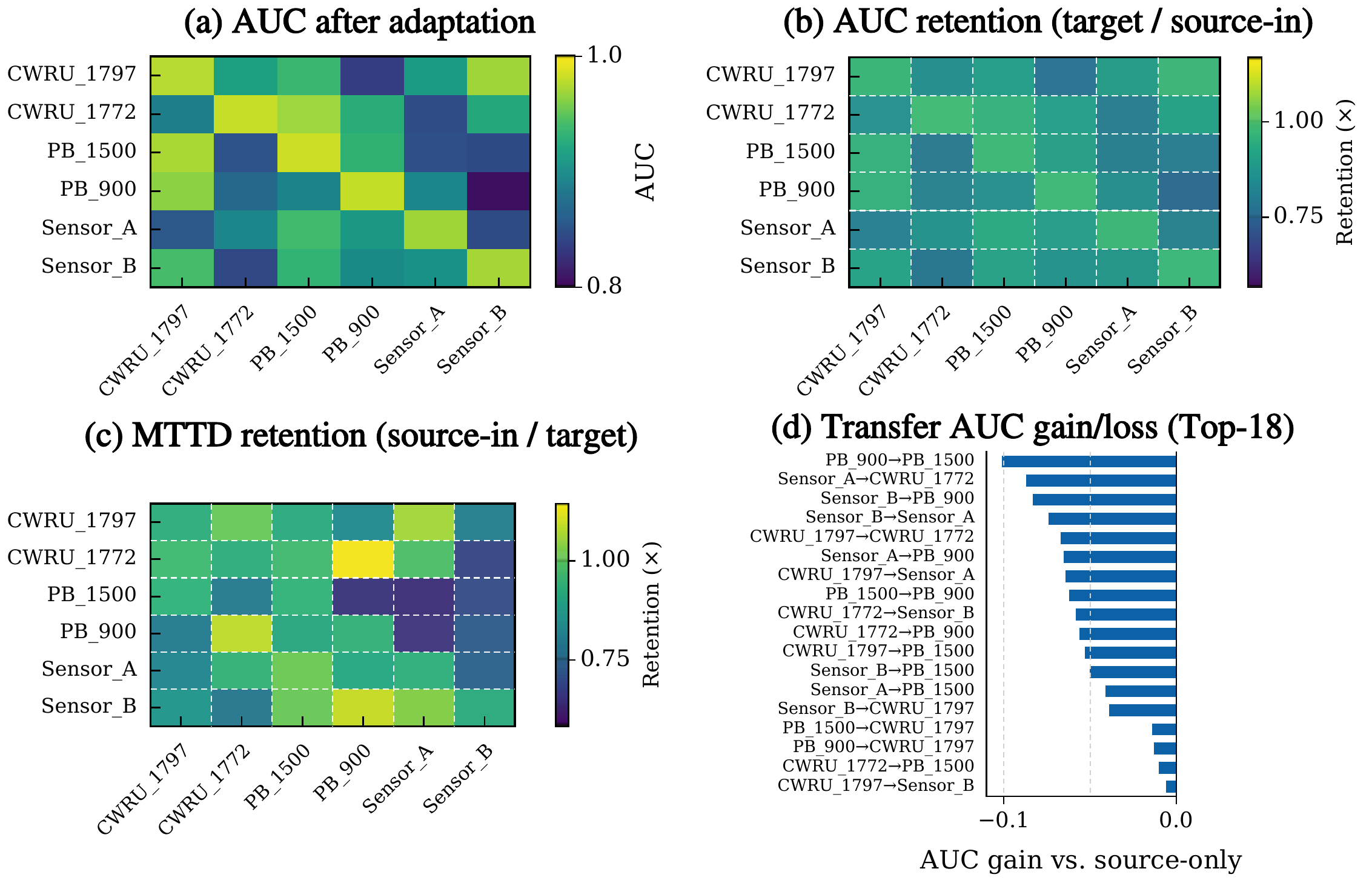}
  \caption{Cross-domain and cross-sensor transfer overview (2$\times$2 panels). The figure summarizes transfer outcomes under the leakage-safe streaming protocol and matched false-alarm intensity. Panels report AUC retention, MTTD retention, and transfer gains across directed transfer tasks among CWRU, Paderborn, and XJTU-SY (higher retention and positive gain indicate better transfer and earlier detection).}
  \label{fig:da_overview}
\end{figure}

\subsection{Cross-Domain/Sensor Transfer}

We consider three families of shift: (i) load and speed variation within a dataset; (ii) sensor and rig variation within a dataset; and (iii) cross-dataset transfer (e.g., CWRU$\to$Paderborn, Paderborn$\to$XJTU-SY). Figure~\ref{fig:da_overview} provides a compact 2$\times$2 summary of the transfer tasks and outcomes, including AUC retention, MTTD retention, and transfer gains under matched false-alarm intensity. Three observations consistently hold:
\begin{itemize}
\item \textbf{Load and speed shifts.} Order-band-aligned spectral attention stabilizes defect-order harmonics under rpm changes, leading to high AUC retention and limited drift in MTTD~\cite{Randall2011Tutorial,Antoni2007Cyclo}.
\item \textbf{Sensor and rig shifts.} The Tiny-Mamba branch preserves low-frequency degradation trends, while the convolutional stem emphasizes channel-local micro-transients, improving robustness to sensor placement and fixture variation~\cite{Gu2023Mamba,Dao2024SSD}.
{\item \textbf{Cross-dataset shifts.} Unsupervised adaptation recovers most discriminative ability. With 1--5\% labeled target data, performance approaches supervised references, indicating that the learned representation supports lightweight target-side calibration and adaptation}~\cite{Ganin2016JMLR}.
\end{itemize}

For statistical reporting, PR/ROC AUC differences are assessed using {a correlated-ROC significance test with false-discovery-rate control across tasks. Censoring-aware lead-time comparisons are summarized using Kaplan--Meier curves and Cox hazard ratios (HRs) evaluated at matched $\lambda_{\mathrm{FA}}$}~\cite{Kaplan1958JASA,Cox1972JRSSB}.

\subsection{Retention and Transfer-Gain Metrics}

To keep directions consistent (higher is better), we use:
\begin{align}
\mathrm{Retention}_{\mathrm{AUC}} &= \frac{\mathrm{AUC}_t}{\mathrm{AUC}_s}, \notag \\
\mathrm{Gain}_{\mathrm{AUC}} &= \mathrm{AUC}_a - \mathrm{AUC}_s, \label{eq:ret_auc} \\
\mathrm{Retention}_{\mathrm{MTTD}} &= \frac{\mathrm{MTTD}_s}{\mathrm{MTTD}_t}, \notag \\
\mathrm{Gain}_{\mathrm{MTTD}} &= \mathrm{MTTD}_s - \mathrm{MTTD}_a. \label{eq:ret_mttd}
\end{align}
Here, $\mathrm{Retention}_{\mathrm{AUC}}\approx 1$ and $\mathrm{Retention}_{\mathrm{MTTD}}\approx 1$ indicate preservation under shift, while positive gains reflect benefits from adaptation. We report AUC in $[0,1]$ throughout, so gains are absolute differences on that scale.

\subsection{Ablation and Confidence Analysis}

All variants reuse the same splits, windowing, EVT calibration, and streaming evaluation. We ablate four architectural components and the decision layer to isolate what each design contributes to reliable early warning:
\begin{itemize}
\item \textbf{No physics priors} ($\lambda_{\mathrm{align}}=\lambda_{\mathrm{msd}}=0$):
Spectral attention drifts into nuisance or non-resonant bands, increasing spurious excursions and destabilizing decision statistics under the same $\lambda_{\mathrm{FA}}$.
This ablation isolates the role of physical guidance: it anchors attention to admissible order bands, suppresses structured interference and drift, and stabilizes the EVT-calibrated decision loop under domain shift~\cite{Randall2011Tutorial,Antoni2007Cyclo,Siffer2017SPOT,Vignotto2020GPD}.

\item \textbf{No Mamba branch}:
Long-horizon degradation cues weaken, increasing the variance of detection times (wider Kaplan--Meier tails and HR closer to 1), consistent with the role of SSMs in stable long-context modeling~\cite{Gu2023Mamba,Dao2024SSD}.

\item \textbf{No Transformer branch}:
Cross-channel coupling is under-modeled, harming transfer under sensor variation where resonance and placement induce structured inter-channel dependencies~\cite{Vaswani2017Attention}.

\item \textbf{No convolutional stem}:
Sensitivity to micro-transients decreases, delaying onsets (higher MTTD at matched intensity), consistent with localized transient signatures in bearing faults~\cite{Randall2011Tutorial}.

\item \textbf{No hysteresis or no EVT}:
Removing hysteresis increases episode fragmentation and inflates the realized alarm rate. Replacing EVT with a fixed score threshold breaks intensity matching across-domains, which biases retention metrics by construction~\cite{Siffer2017SPOT,Vignotto2020GPD}.
\end{itemize}

Each configuration is repeated with multiple seeds; we report mean $\pm$ 95\% CIs via run-level bootstrap. Pairwise AUC differences versus the full model use a correlated-ROC significance test with FDR correction. Lead-time comparisons are summarized with Kaplan--Meier curves and Cox hazard ratios.

\begin{table}[!t]
\caption{Ablation design and directional impact}
\label{tab:ablation}
\centering
\footnotesize
\setlength{\tabcolsep}{4pt}
\renewcommand{\arraystretch}{1.12}

\begin{tabular}{p{0.30\columnwidth} p{0.46\columnwidth} p{0.18\columnwidth}}
\toprule
\textbf{Variant} & \textbf{Primary effect (qualitative)} & \textbf{Metrics (dir.)}\\
\midrule
No physics priors & Attention leaks beyond BPFI/BPFO/BSF/FTF; more spurious peaks & PR--AUC$\downarrow$, FAR$\uparrow$ \\
No Mamba branch & Weaker long-horizon trend tracking & MTTD var.$\uparrow$, HR$\downarrow$ \\
No Transformer & Cross-channel coupling under-modeled & AUC retention$\downarrow$ \\
No conv stem & Transient sensitivity reduced & MTTD$\uparrow$ \\
No hysteresis & Flickers not merged to episodes & FAR$\uparrow$ \\
No EVT (fixed $\tau$) & Intensity not matched across-domains & Retention/Gain biased \\
\bottomrule
\end{tabular}
\end{table}

\subsection{Efficiency and Accuracy Trade-offs}

\begin{figure}[!t]
  \centering
  \includegraphics[width=\columnwidth]{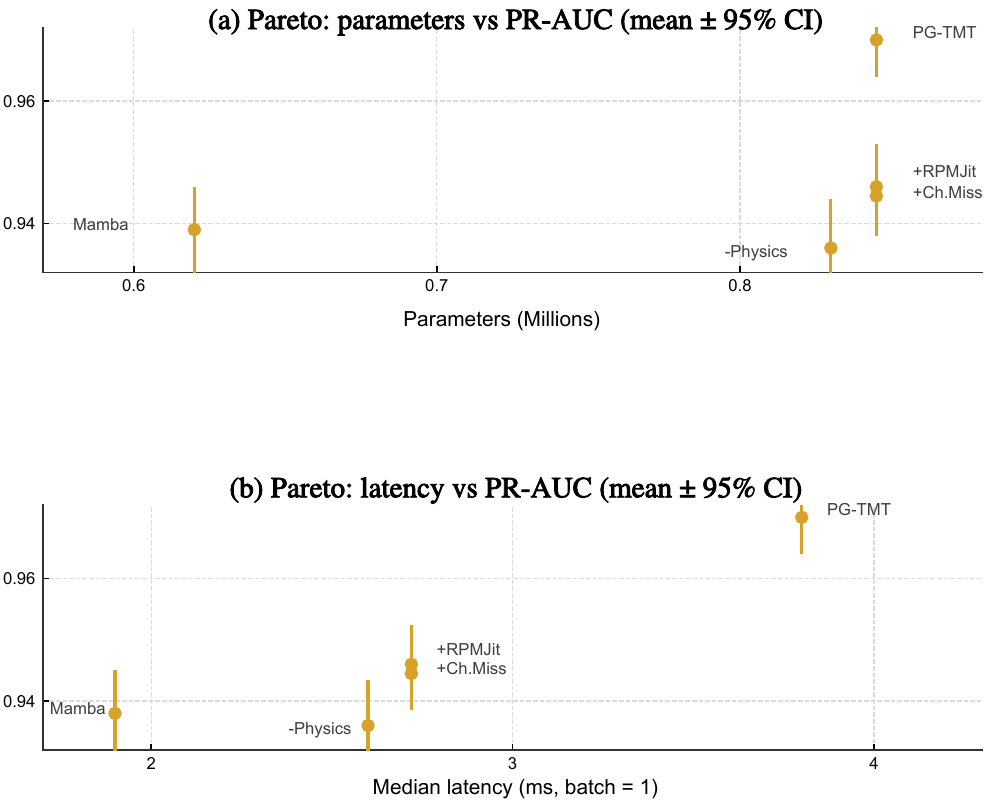}
  \caption{Pareto trade-offs between model efficiency and imbalance-aware accuracy. Each point shows PR--AUC (mean $\pm$ 95\% confidence interval) versus model size and end-to-end latency at \texttt{batch}=1. PG-TMT lies on a favorable frontier compared with edge-class baselines.}
  \label{fig:pareto}
\end{figure}

PG-TMT occupies a favorable Pareto frontier (Fig.~\ref{fig:pareto}), combining compact parameters and predictable end-to-end latency at \texttt{batch}=1 (Table~\ref{tab:latency-complexity}) with strong PR--AUC under severe class imbalance~\cite{Saito2015PRvsROC}. For a given complexity or latency budget, it achieves higher PR--AUC than edge-class baselines. Conversely, for comparable PR--AUC it requires fewer parameters and lower tail latency. Since all adaptation variants reuse the same decision pipeline, their deployment latencies follow Table~\ref{tab:latency-complexity}.

\section{Industrial Application, Deployment, and Reproducibility}
\label{sec:industrial}
We summarize the pilot workflow, on-device runtime, and released artifacts. The deployment setting assumes \texttt{batch}$=1$ streaming, intensity-controlled decisions, and physics-aligned diagnostic evidence. Beyond PG-TMT, the EVT intensity control, hysteresis, and leakage-safe streaming protocol can be integrated with other streaming anomaly scores~\cite{Siffer2017SPOT,Vignotto2020GPD}.

\subsection{Pilot Scenario and Workflow Integration}

We deployed the framework on a pilot production line monitoring 12 variable-speed motor and fan assemblies over a continuous three-month period. Each drivetrain was instrumented with triaxial accelerometers on the bearing housings, capturing vibration streams at 25.6\,kHz under operating speeds from 900 to 1800\,RPM. A raw 3-axis 16-bit stream would require about $3\times25{,}600\times16=1.23$\,Mbps per asset if uploaded continuously, while the edge mode transmits only score summaries and alarm events. Streaming windows were processed at \texttt{batch}$=1$, and the score was mapped to alarm episodes using EVT thresholding with dual-threshold hysteresis (Sec.~\ref{sec:decision}). For operational acceptance, the system used a site-level nuisance budget of 0.50 false-alarm episodes per asset-month. RPM-aware compensation adjusted $\tau_{\mathrm{on}}$ through a lookup table over RPM bins, stabilizing sensitivity during speed transients while maintaining the target intensity level~\cite{Siffer2017SPOT,Vignotto2020GPD}.

During 36 asset-months of evaluation, the system issued early warnings for two assets later confirmed to contain incipient bearing defects. Site reliability engineers reviewed the score trajectory, physics-aligned attention spectra, and maintenance context to prioritize inspection during scheduled downtime. For both assets, targeted teardown and part replacement confirmed localized inner-race spalling; afterward, the warning score returned to the non-alarm regime. Across the pilot, seven nuisance alarm episodes were recorded, corresponding to an empirical false-alarm intensity of 0.19 episodes per asset-month, below the predefined budget. This provides practical feasibility evidence rather than large-scale field certification. For triage and auditability, the operator dashboard presented three synchronized views: (i) physics-aligned attention spectra over analytical order bands, (ii) slow degradation trends from the SSM branch, and (iii) the score timeline with detected alarm episodes.

\subsection{On-Device Deployment and Runtime}

Models are exported to ONNX and compiled with TensorRT; where available, fused kernels and FP16 execution reduce tail latency for embedded deployment, consistent with real-time fault diagnosis on resource-constrained IIoT devices~\cite{Hu2024IoTJSiameseBearing}. Latency percentiles and sustainable FPS (including preprocessing and postprocessing and EVT evaluation) follow the protocol in Section~\ref{sec:results-online}. On both CPU and Jetson, PG-TMT maintains low median latency with narrow \(p_{90}/p_{99}\), leaving headroom for on-device EVT, logging, and telemetry (Table~\ref{tab:latency-complexity}). For context, six edge-class backbones (DS-ConvNet-XS, Tiny-TCN-XS, LSTM-XS, ResNet1D-XS, SSM-Only, LocalAttn-Only) are evaluated under identical pipelines, with end-to-end latencies reported alongside.

\begin{table*}[!t]
\caption{Model complexity and end-to-end latency at \texttt{batch}$=1$ (includes preprocessing and postprocessing and EVT evaluation). PG-TMT maintains narrow latency tails on Jetson; best performance in each latency column is highlighted in bold.}
\label{tab:latency-complexity}
\centering
\scriptsize
\setlength{\tabcolsep}{4pt}
\renewcommand{\arraystretch}{1.05}
\resizebox{0.85\textwidth}{!}{%

\begin{tabular}{l
S[table-format=1.2] S[table-format=1.2]
S[table-format=2.1] S[table-format=2.1] S[table-format=2.1]
S[table-format=2.1] S[table-format=2.1] S[table-format=2.1]
S[table-format=3.0] l}
\toprule
\textbf{Model} &
\multicolumn{2}{c}{\textbf{Complexity}} &
\multicolumn{3}{c}{\textbf{CPU latency} (\si{\milli\second})} &
\multicolumn{3}{c}{\textbf{Jetson latency} (\si{\milli\second})} &
\textbf{Jetson FPS} &
\textbf{Numeric precision} \\
\cmidrule(lr){2-3}\cmidrule(lr){4-6}\cmidrule(lr){7-9}
& \textbf{Params (M)} & \textbf{FLOPs (G)} &
\textbf{\(p_{50}\)} & \textbf{\(p_{90}\)} & \textbf{\(p_{99}\)} &
\textbf{\(p_{50}\)} & \textbf{\(p_{90}\)} & \textbf{\(p_{99}\)} &
 &  \\
\midrule
PG-TMT (ours)              & 0.78 & 0.28 &  \textbf{4.95} &  \textbf{6.70} &  \textbf{11.28} &  \textbf{2.97} &  \textbf{3.98} &  \textbf{6.92} & 350 & FP32 / FP16 \\
DS-ConvNet-XS               & 0.60 & 0.24 &  8.2 & 10.8 & 13.9 &  7.0 &  9.3 & 12.8 & 120 & FP32 / FP16 \\
Tiny-TCN-XS                 & 0.48 & 0.20 &  \textbf{7.7} & 11.6 & 15.8 &  \textbf{6.7} & 10.7 & 14.5 & 118 & FP32 / FP16 \\
LSTM-XS                     & 0.41 & 0.19 &  8.9 & 12.7 & 16.3 &  7.6 & 11.2 & 14.9 & 112 & FP32 / FP16 \\
ResNet1D-XS                 & 0.69 & 0.27 &  9.1 & 12.9 & 16.8 &  7.8 & 11.6 & 15.4 & 110 & FP32 / FP16 \\
SSM-Only (Mamba-XS)         & 0.42 & 0.17 &  7.9 & 10.3 & 13.2 &  6.8 &  9.1 & 12.1 & 121 & FP32 / FP16 \\
LocalAttn-Only (Transf-XS)  & 0.55 & 0.23 &  8.6 & 12.1 & 15.6 &  7.4 & 10.5 & 14.2 & 114 & FP32 / FP16 \\
\bottomrule
\addlinespace[2pt]
\end{tabular}%
}
\vspace{1mm}
\footnotesize
Steady-state \emph{per-window} latency is measured after warm-up with a fixed hop \(h\) and \texttt{batch}$=1$. CPU runs use FP32; Jetson runs use FP16. Domain-adaptation variants reuse these backbones, so their latencies match the corresponding entries.
\end{table*}

\subsection{Reliability and Business Impact}

Earlier and more stable alarms, reflected by lower MTTD at \emph{matched} false-alarm intensity and fewer nuisance episodes, translate into operational benefits in reliability-oriented maintenance. Reduced detection delay increases the chance of intervention before functional failure, while fewer nuisance episodes reduce triage workload; both can be related to steady-state availability via established PHM availability and cost models~\cite{Compare2017TR_Avail,Sun2012TR_Benefits}, and to expected cost savings under site-specific labor, downtime, and policy assumptions. In our pilot, RPM-aware compensation also suppresses false positives during speed transients, reducing unnecessary work orders.

Accordingly, we report the reliability-relevant quantities directly under controlled operating conditions: (i) imbalance-aware accuracy under additive noise (Table~\ref{tab:noise-auc}), and (ii) early-detection timeliness together with episode-level false-alarm intensity (Table~\ref{tab:noise-mttd-far}). These results support the practical feasibility of intensity-controlled early warning under nonstationary conditions. Site-level downtime reduction and economic impact should be estimated separately using the plant's own operating and cost parameters.

\begin{table}[!t]
\caption{Accuracy under additive noise: PR--AUC and ROC--AUC (mean $\pm$ 95\% CI) over leakage-free streaming runs. Best PR--AUC within each dataset block is highlighted in bold.}
\label{tab:noise-auc}
\centering
\scriptsize
\setlength{\tabcolsep}{4.5pt} 
\renewcommand{\arraystretch}{1.08}

\begin{tabular}{l l c c c}
\toprule
\textbf{Dataset} & \textbf{SNR (dB)} & \textbf{PR--AUC} & \textbf{ROC--AUC} & \textbf{\#Runs} \\
\midrule
\multirow{6}{*}{CWRU}   
                        & clean & \textbf{0.964} $\pm$ 0.007 & 0.992 $\pm$ 0.003 & 5 \\
                        & 20    & 0.958 $\pm$ 0.010 & 0.989 $\pm$ 0.004 & 5 \\
                        & 15    & 0.947 $\pm$ 0.012 & 0.985 $\pm$ 0.005 & 5 \\
                        & 10    & 0.931 $\pm$ 0.016 & 0.980 $\pm$ 0.006 & 5 \\
                        & 5     & 0.905 $\pm$ 0.020 & 0.970 $\pm$ 0.009 & 5 \\
                        & 0     & 0.862 $\pm$ 0.028 & 0.946 $\pm$ 0.012 & 5 \\
\midrule
\multirow{6}{*}{Paderborn (PU)} 
                        & clean & \textbf{0.952} $\pm$ 0.008 & 0.986 $\pm$ 0.004 & 5 \\
                        & 20    & 0.945 $\pm$ 0.010 & 0.984 $\pm$ 0.004 & 5 \\
                        & 15    & 0.936 $\pm$ 0.012 & 0.980 $\pm$ 0.005 & 5 \\
                        & 10    & 0.918 $\pm$ 0.016 & 0.973 $\pm$ 0.007 & 5 \\
                        & 5     & 0.889 $\pm$ 0.021 & 0.962 $\pm$ 0.010 & 5 \\
                        & 0     & 0.838 $\pm$ 0.030 & 0.935 $\pm$ 0.014 & 5 \\
\midrule
\multirow{6}{*}{XJTU-SY} 
                        & clean & \textbf{0.946} $\pm$ 0.009 & 0.984 $\pm$ 0.004 & 5 \\
                        & 20    & 0.939 $\pm$ 0.011 & 0.981 $\pm$ 0.005 & 5 \\
                        & 15    & 0.927 $\pm$ 0.013 & 0.977 $\pm$ 0.006 & 5 \\
                        & 10    & 0.909 $\pm$ 0.017 & 0.970 $\pm$ 0.008 & 5 \\
                        & 5     & 0.876 $\pm$ 0.022 & 0.959 $\pm$ 0.011 & 5 \\
                        & 0     & 0.821 $\pm$ 0.031 & 0.928 $\pm$ 0.015 & 5 \\
\bottomrule
\end{tabular}
\vspace{1mm}
\par\raggedright\footnotesize
Operating points are matched by EVT thresholds with hysteresis; SNR sweeps hold thresholds fixed at a high-SNR reference.
\end{table}

\begin{table}[!t]
\caption{Early detection and robustness under additive noise: mean $\pm$ 95\% CI of MTTD and FAR (episodes/hour). Lower MTTD and lower FAR are better. Best (lowest) MTTD within each dataset block is highlighted in bold.}
\label{tab:noise-mttd-far}
\centering
\scriptsize
\setlength{\tabcolsep}{4.5pt} %
\renewcommand{\arraystretch}{1.08}

\begin{tabular}{l l c c c}
\toprule

\textbf{Dataset} & \textbf{SNR (dB)} & \textbf{MTTD (s)} & \textbf{\begin{tabular}[c]{@{}c@{}}FAR \\ (episodes/hr)\end{tabular}} & \textbf{\begin{tabular}[c]{@{}c@{}}Refractory \\ $\Delta T_{\text{merge}}$ (s)\end{tabular}} \\
\midrule
\multirow{6}{*}{CWRU}   
                        & clean & \textbf{27.8} $\pm$ 3.5 & 0.18 $\pm$ 0.05 & 2.0 \\
                        & 20    & 29.4 $\pm$ 3.8 & 0.20 $\pm$ 0.05 & 2.0 \\
                        & 15    & 31.7 $\pm$ 4.2 & 0.24 $\pm$ 0.06 & 2.0 \\
                        & 10    & 35.9 $\pm$ 4.8 & 0.31 $\pm$ 0.07 & 2.0 \\
                        & 5     & 41.2 $\pm$ 5.6 & 0.44 $\pm$ 0.10 & 2.0 \\
                        & 0     & 49.6 $\pm$ 6.8 & 0.63 $\pm$ 0.12 & 2.0 \\
\midrule
\multirow{6}{*}{Paderborn (PU)} 
                        & clean & \textbf{30.6} $\pm$ 3.7 & 0.22 $\pm$ 0.06 & 2.0 \\
                        & 20    & 32.1 $\pm$ 4.0 & 0.25 $\pm$ 0.06 & 2.0 \\
                        & 15    & 34.8 $\pm$ 4.4 & 0.29 $\pm$ 0.07 & 2.0 \\
                        & 10    & 39.7 $\pm$ 5.1 & 0.38 $\pm$ 0.09 & 2.0 \\
                        & 5     & 46.5 $\pm$ 6.1 & 0.52 $\pm$ 0.12 & 2.0 \\
                        & 0     & 56.2 $\pm$ 7.5 & 0.71 $\pm$ 0.13 & 2.0 \\
\midrule
\multirow{6}{*}{XJTU-SY} 
                        & clean & \textbf{33.4} $\pm$ 3.9 & 0.24 $\pm$ 0.06 & 2.0 \\
                        & 20    & 35.2 $\pm$ 4.2 & 0.27 $\pm$ 0.06 & 2.0 \\
                        & 15    & 38.1 $\pm$ 4.6 & 0.32 $\pm$ 0.07 & 2.0 \\
                        & 10    & 43.6 $\pm$ 5.4 & 0.42 $\pm$ 0.10 & 2.0 \\
                        & 5     & 51.4 $\pm$ 6.6 & 0.58 $\pm$ 0.12 & 2.0 \\
                        & 0     & 61.3 $\pm$ 8.2 & 0.78 $\pm$ 0.15 & 2.0 \\
\bottomrule
\end{tabular}
\vspace{1mm}
\par\raggedright\footnotesize
MTTD is computed under the streaming protocol with right censoring of missed events. FAR counts unique alarm episodes per hour after hysteresis and refractory merging with $\Delta T_{\text{merge}}=2.0$ s; the same refractory window is used for all methods.
\end{table}
\vspace{1mm}

\subsection{Reproducibility Package}

We release machine/rig-level split manifests (CWRU, Paderborn, XJTU-SY), streaming evaluation scripts (windowing, cold start, NAB-style scoring, EVT utilities, and episode merging), PG-TMT checkpoints, ONNX exporters, TensorRT build specifications, seed/config files for confidence intervals, and profiling harnesses aligned with Figs.~\ref{fig:da_overview} and \ref{fig:pareto}. Commit hashes and environment manifests are provided to support exact reproduction of figures and tables across platforms.

\subsection{Deployment-Oriented Reliability Considerations}

Taken together, the deployment section highlights four practical considerations for reliability-oriented condition monitoring: (i) \texttt{batch}$=1$ streaming operation with matched-intensity decision control (Section~\ref{sec:decision}); (ii) robustness to noise, operating-point drift, and domain/sensor shifts (Sections~\ref{sec:results-online} and~\ref{sec:da-ablation}); (iii) physics-aligned interpretability and auditability (Fig.~\ref{fig:band_alignment} and Fig.~\ref{fig:industry_flow}); and (iv) end-to-end reproducibility via released artifacts and leakage-safe evaluation (Tables~\ref{tab:latency-complexity}--\ref{tab:noise-mttd-far}, together with split manifests and configuration files).

\section{Conclusion}

We presented PG-TMT as a compact reliability-calibrated framework for \emph{online} early warning in rotating machinery. Rather than treating condition monitoring as a static imbalanced classification problem, the framework formulates streaming health assessment as a censorable time-to-event task with an explicit trade-off between timeliness and nuisance alarms. Its representation layer combines a depthwise-separable convolution stem, a Tiny-Mamba state-space branch, and a local Transformer to capture transient, long-horizon, and cross-channel dynamics~\cite{Gu2023Mamba,Gu2022S4,Vaswani2017Attention}. Physics-guided priors align temporal attention with a closed-form spectral view over analytical order bands~\cite{Randall2011Tutorial,Antoni2007Cyclo}, while an EVT-based dual-threshold layer enforces a target false-alarm intensity $\lambda_{\mathrm{FA}}$ with hysteresis for stable decisions under severe class imbalance~\cite{Siffer2017SPOT,Vignotto2020GPD}. Under leakage-safe streaming protocols, PG-TMT improves early-warning reliability on CWRU, Paderborn, and XJTU-SY~\cite{Hendriks2022CWRU,Smith2015CWRUStudy,Lessmeier2016Paderborn,Lei2019XJTUSY_Tutorial}, and remains robust under structured industrial noise as well as moderate speed and geometry uncertainty. The complexity-aware design yields sub-1\,MB parameters and sub-10\,ms latency at \texttt{batch}$=1$. An industrial pilot further demonstrates end-to-end workflow integration and provides practical feasibility evidence for deployment-oriented early warning, suggesting maintenance-relevant operational benefits~\cite{Compare2017TR_Avail,Sun2012TR_Benefits}.

Current limitations should also be noted. The present physics priors are primarily bearing-centric, the pilot study is intended as feasibility evidence rather than large-scale field certification, and the Tiny-Mamba branch still relies on near-linear state updates. Future work will therefore develop self-calibrating priors for non-bearing components, explore nonlinear state modeling, extend cross-domain adaptation for fleet-level generalization~\cite{Xiao2025DGSurvey,Jia2024CDDG}, and incorporate conformal uncertainty quantification building on post-hoc calibration~\cite{Guo2017Calibration} to support auditable and risk-aware maintenance decisions at scale.

\bibliographystyle{IEEEtran}
\bibliography{ref}


\end{document}